\documentclass[a4paper,fleqn]{cas-dc}
\usepackage{framed}
\usepackage{multirow}
\usepackage[switch]{lineno}
\usepackage[utf8]{inputenc}
\usepackage{algorithm}
\usepackage{algpseudocode}
\usepackage{lineno}         %
\usepackage{gensymb}
\usepackage{textcomp}
\usepackage{amsmath}
\usepackage{graphicx}
\usepackage{amsfonts}
\usepackage{booktabs}
\usepackage{subcaption}
\usepackage{soul}
\usepackage[numbers]{natbib}
\usepackage{float}
\usepackage{comment}

\begin{document}
\let\WriteBookmarks\relax
\def\floatpagepagefraction{1}
\def\textpagefraction{.001}
% \linenumbers

\shorttitle{Adapted CHCI for Non-Euclidean TSPs}
\shortauthors{M. Goutham et~al.}

% \title [mode = title]{Adapted Convex Hull Cheapest Insertion Heuristic: an Efficient Tour Construction Heuristic for Non-Euclidean TSPs}        

\title [mode = title]{
A convex hull cheapest insertion heuristic for the non-Euclidean TSP
}
\nonumnote{DOI: \url{https://doi.org/10.1016/j.robot.2026.105685}}

\author[1]{Mithun Goutham}[orcid=0000-0001-5789-7590]\ead[style=plain]{mithun9@uw.edu}
\author[2]{Ethan David Rosati}
\author[3]{Hollis Schuler}
\author[4]{Meghna Menon}
\author[4]{Sarah Garrow}
\author[2]{Stephanie Stockar}

\affiliation[1]{{University of Washington, Seattle, WA 98195, USA}}
\affiliation[2]{{The Ohio State University, Columbus, OH 43210 USA}}
\affiliation[3]{{University of Colorado, Boulder, CO 80309 USA}}
\affiliation[4]{{Ford Motor Company, Dearborn, MI 48126, USA}}

\begin{abstract}
Autonomous robots frequently encounter routing problems that involve non-Euclidean cost considerations due to obstacles, traffic, or a cost function that is not simply the straight-line distance between locations to be visited.
Often, the resulting Non-Euclidean Traveling Salesperson Problem (NETSP) must be solved onboard with limited computational resources, posing a significant challenge due to its NP-hard combinatorial nature.
To address this, the Adapted Convex Hull Cheapest Insertion (ACHCI) algorithm is proposed. 
% ACHCI is a lightweight heuristic specifically designed to compute NETSP tours in real-time using the limited memory budgets and processing speeds typical of small form factor robots.
\textcolor{black}{
ACHCI is a lightweight heuristic designed for resource-constrained onboard tour computation, with small form factor robots as its target application.}
ACHCI combines a multidimensional scaling approach with a convex hull initialized tour construction procedure to generalize the well-known Euclidean CHCI heuristic to non-Euclidean problems.
Computational experiments on diverse modified TSPLIB scenarios demonstrate that ACHCI outperforms other lightweight heuristics like Nearest Neighbor and Nearest Insertion in 88\% and 99\% of the cases, as well as population-based metaheuristics such as Genetic Algorithms and Ant Colony Optimization in 87\% and 95\% of test cases respectively.
% By enabling real-time routing under the limited computational resources available, ACHCI enhances the operational efficiency of autonomous agents by reducing travel distance, energy consumption, charging-related downtime, task completion duration and operating costs.
\textcolor{black}{
The adoption of ACHCI for resource-limited onboard routing is expected to} enhance the operational efficiency of autonomous agents by reducing travel distance, energy consumption, charging-related downtime, task completion duration and operating costs.
\end{abstract}

\begin{keywords}
Traveling salesman problems \sep 
Vehicle routing and navigation \sep 
Control and Scheduling \sep 
Algorithms \sep 
Logistics \sep 
Supply Chains \sep 
\end{keywords}

\maketitle

\section{Introduction}\label{sec:intro}

Many logistics and inspection operations give rise to the Non-Euclidean Traveling Salesperson Problem (NETSP), where an agent must visit a set of locations while minimizing travel costs affected by environmental and operational constraints \cite{boyaci2021vehicle}.
For example, this routing problem is increasingly relevant in warehouses and flexible manufacturing systems, where autonomous mobile robots must retrieve multiple items from locations across dense facility layouts, navigating aisles, one-way corridors, and restricted zones \cite{fragapane2021planning}.
Similar challenges arise in disaster response operations or electricity grid inspections, where drones must visit locations spread across wind or solar farms, mountainous terrains, or no-fly zones \cite{guerrero2013uav}.
Likewise, waste collection, courier or food delivery agents must navigate the urban road network with dynamic traffic conditions that affect route selection \cite{papalitsas2019unconventional}.

\textcolor{black}{
Existing approaches to routing in obstacle-laden environments, such as visibility graph methods and spatial partitioning techniques, compute pairwise travel costs via shortest-path algorithms on visibility graphs first, and then determine a visit ordering over these costs as a second step \cite{alkema2022tsp, arora1998polynomial, mitchell1999guillotine, kisfaludi2025realizing}.
These approaches require explicit waypoint coordinates and a polygonal description of obstacles in the environment to construct the underlying geometric graph used for shortest-path computation. 
In practical large-scale deployments, the path-cost computation is performed by a dedicated path planner on a central computer \cite{saha2006planning, goutham2023optimal, danner2000randomized}.}

With the increasing adoption of edge computing and decentralized control strategies, deployed agents are expected to autonomously compute their route, using only their onboard processing capabilities \cite{lee2022uav,guo2024decentralized,xia2023survey}. 
\textcolor{black}{
Therefore, this paper focuses on the second step in routing, where, given a precomputed non-Euclidean cost matrix, the goal is to determine a low-cost visit ordering 
\textcolor{black}{
using an algorithm designed for onboard execution}
under the limited memory constraints and processing capabilities of microcontroller-based autonomous agents.}

\textcolor{black}{Using exact algorithms to find} the optimal solution to the NETSP, \textcolor{black}{even after the costs have been computed}, is challenging because the routing problem is NP-hard and requires significant computational capabilities, even for a moderate number of locations \cite{saalweachter2008non,karp2010reducibility,shobaki2015exact,salii2019revisiting,goutham2024novel}.
\textcolor{black}{
Heuristics, on the other hand, generate feasible solutions quickly but they do not guarantee optimality, and for this reason, heuristics are typically used when solutions have to be found instantaneously \citep{wong2014dynamic, markovic2015optimizing}, or to initialize exact methods for faster convergence to the optimal solution \citep{braekers2014exact}.
}

\textcolor{black}{
Tour improvement heuristics iteratively refine an initial candidate tour by exploring similar tours by local modifications, yielding near-optimal solutions without exhaustive search \cite{aarts2018local,rego2011traveling}.
These methods range from single-solution improvement approaches to population-based metaheuristics that evolve multiple candidate tours simultaneously.
The Genetic Algorithm (GA) is one such population-based metaheuristic, which, when applied to the NETSP, iteratively improves a population of candidate tours, arriving at a low cost tour \cite{larranaga1999genetic}.
The GA does not explicitly assign costs to individual edges that constitute a tour and instead modifies tours through selection, crossover, and mutation operations inspired by natural evolution.
In contrast, the Ant Colony Optimization (ACO) algorithm improves candidate tours by explicitly assigning values to edges, using these to probabilistically arrive at lower-cost tours \cite{dorigo1996ant}.
Both GA \cite{mininno2008real,dendaluce2014microcontroller,griffiths1998microcontroller,mamdoohi2012implementation} and ACO \cite{dahmane2020ant,scheuermann2004fpga,ilhan2016solution} have been implemented on embedded hardware platforms, making them relevant benchmarks for resource-constrained onboard deployment contexts targeted in this work.
}

\textcolor{black}{
The Lin-Kernighan-Helsgaun (LKH) algorithm is one of the most sophisticated tour improvement heuristics, removing k edges from a candidate tour and reconnecting the tour in a better order, often} yielding near-optimal results \cite{lin1973effective,helsgaun2000effective}.
\textcolor{black}{However, unlike GA  and ACO, no equivalent microcontroller or embedded system deployment of the LKH algorithm has been reported in the literature.
In fact, the only reported on-board deployment of LKH} required systems with at least 4 GB of RAM and 1.6 GHz processors running desktop-class operating systems \cite{povzgaj2024overview,hernandez2017multi}.
\textcolor{black}{
As a result, sophisticated algorithms like LKH} have not yet been adopted in decentralized control strategies with onboard computing, where the microcontrollers commonly found in small form factor robots operate with memory budgets and processing speeds that are three to four orders of magnitude smaller than the required capability \cite{kakoty2024microcontrollers}.

Tour construction heuristics are a more computationally efficient approach, building tours by incrementally selecting the next location to visit, leveraging spatial proximity or convex hull boundaries, and have been extensively developed in the Euclidean context, where the cost between locations is simply their straight-line distance, permitting the geometry-aware exploitation of location coordinates \cite{huang2017investigating}.
\textcolor{black}{
A promising approach is the Christofides algorithm, a sophisticated tour construction heuristic that combines a minimum spanning tree with a minimum-weight perfect matching (MWPM) step to guarantee a tour cost no more than 1.5 times optimal in metric TSPs \cite{christofides2022worst}.
However, the Christofides algorithm has not been demonstrated on onboard embedded or microcontroller-based systems in existing literature, likely due to the implementation demands of its MWPM step which relies on dynamic graph structures and nontrivial memory management \cite{kolmogorov2009blossom,kakoty2024microcontrollers}.
%dynamic memory allocation, recursive blossom contraction, and auxiliary graph structures \cite{kolmogorov2009blossom} — 
The standard approach, similar to LKH, is central computation  on desktop-class hardware with the resulting tour path communicated to the robot for onboard execution \cite{mandal2021planning}.
}

In addition to defining a low cost NETSP tour, the onboard computer must also accomplish other critical onboard tasks such as localization, sensor fusion, and control.
These requirements motivate the utilization of lightweight heuristics that reduce computational demands by exploiting problem structure and geometric approximations \cite{kim2017brain}.
\textcolor{black}{
For this reason, this paper focuses on developing a lightweight, real-time compatible tour construction heuristic specifically tailored for the NETSP,}
\textcolor{black}{with resource-constrained embedded and onboard computation as a target application.}

Among lightweight heuristics, the Convex Hull Cheapest Insertion (CHCI) heuristic has been shown to produce superior solutions compared to greedy local search methods such as the Nearest Neighbor (NN) or Nearest Insertion (NI) heuristics \cite{ivanova2021methods,huang2017investigating}.
The CHCI heuristic is initiated by a sub-tour of the convex hull of locations, and interior points are progressively incorporated into the sub-tour in increasing order of insertion cost, until every location is visited.
This strategy is advantageous because convex hull boundary locations are visited in the same cyclic order as they appear in the optimal Euclidean TSP tour \cite{deineko1994convex,golden1980approximate}.

In the NETSP context, however, the spatial arrangement of the locations no longer directly defines the cost, and consequently, effective tour construction heuristics tailored for the NETSP have been neglected in literature despite their speed and low computational demands \cite{glover2001construction}.
As a result, the promising CHCI heuristic has never been adapted to the non-Euclidean problem because its initialization depends on constructing a convex hull, which is unavailable in non-Euclidean problems \cite{huang2017investigating}.
Instead, NN or NI heuristics are used, or the non-Euclidean cost function is substituted with straight-line distances \cite{alkema2022tsp, faigl2011application}.
However, these approximations do not account for the added constraints of NETSPs, resulting in sub-optimal solutions, especially when the true costs deviate significantly from the Euclidean distance. 

This paper extends the Euclidean CHCI algorithm to the non-Euclidean setting, motivated by the expected reduction in tour cost when compared to the commonly used NN and NI heuristics that rely solely on local cost information.
The novel Adapted CHCI (ACHCI) heuristic achieves this by first applying multidimensional scaling (MDS) \textcolor{black}{to embed the non-Euclidean cost matrix into a projected Euclidean space}, finding a set of points whose pairwise Euclidean distances approximate the non-Euclidean pairwise costs \cite{torgerson1952multidimensional, crippen1978note}.
The points on the boundary of their convex hull are used to initiate a sub-tour, onto which the remaining points are added in order of their true non-Euclidean insertion cost ratios.
While MDS has recently been applied to the NETSP for tour length estimation, local clustering, and to understand human cognition \cite{kou2022optimal,huang2018efficient, vandrunen2022traveling}, the use of MDS to initiate the CHCI heuristic for the NETSP is a novel approach.
\begin{figure*}[b]
    \centering
    \subfloat[Warehouse layout with multiple visit locations\label{fig:warehouse}]{%
        \includegraphics[trim =0mm 60mm 128mm 0mm, clip, width=0.48\linewidth]{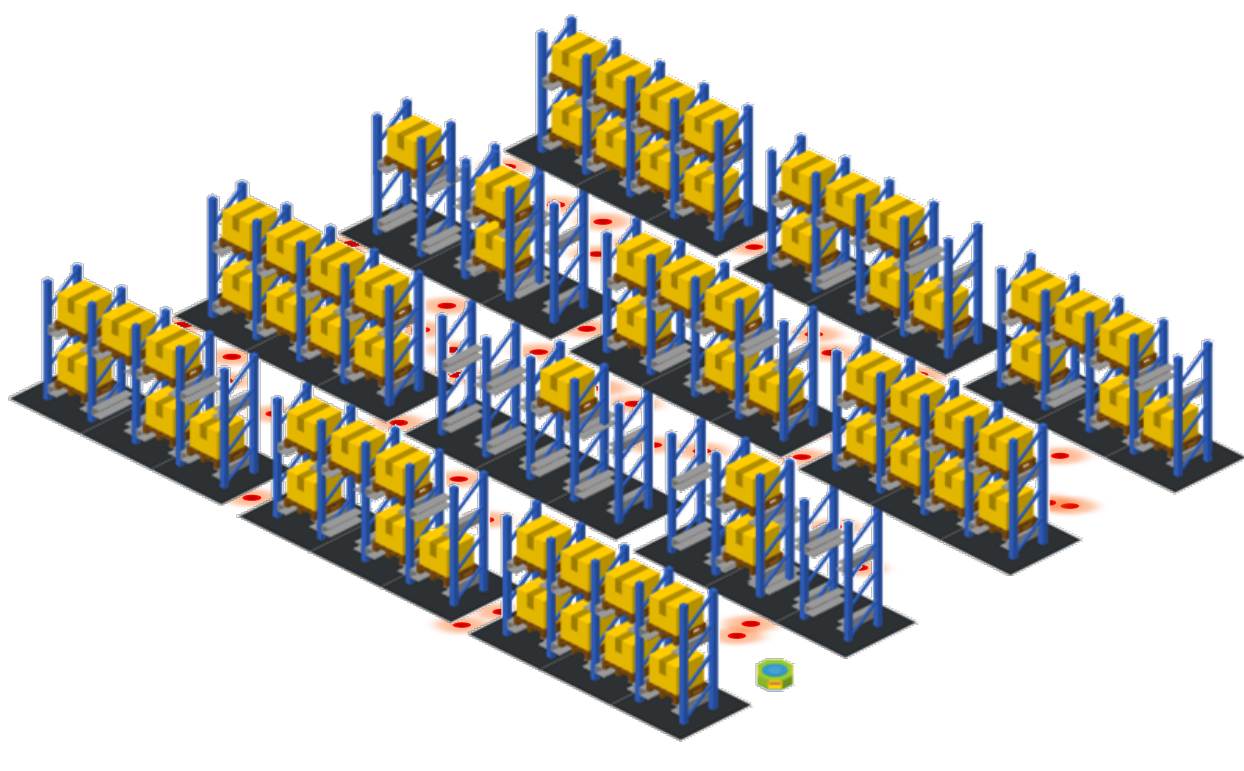}
    } \hfill
    \subfloat[Initial sub-tour \label{fig:initialTour}]{%
        \includegraphics[trim = 89.2mm 96mm 85mm 92mm, clip, width=0.17\linewidth]{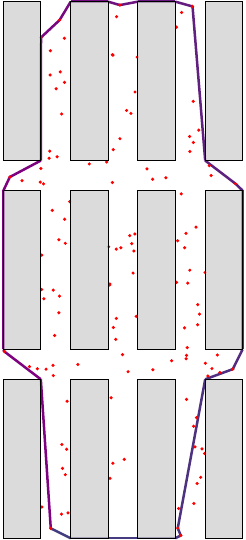}
    } \hfill
    \subfloat[TSP solution\label{fig:TSPNEsolution}]{%
        \includegraphics[trim =89.2mm 96mm 85mm 92mm, clip, width=0.17\linewidth]{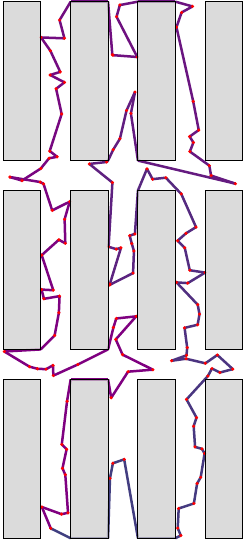}
    }
    \caption{Non-Euclidean TSP to be solved by an autonomous robot within a warehouse facility}
    \label{fig:TSPNEInstance}
\end{figure*}

% The remainder of this paper is organized as follows. 
% Section 2 mathematically formulates the NETSP, after which the proposed ACHCI algorithm is defined and explained using an illustrative example in Section 3. Section 4 defines the benchmark algorithms used for comparison, and Section 5 presents computational results comparing ACHCI against these benchmarks across 57 modified TSPLIB instances that reflect a broad range of real-world NETSP scenarios \cite{reinhelt2014tsplib}.

The remainder of this paper is organized as follows. 
First, the NETSP is defined in Section \ref{NETSP}, after which the proposed ACHCI algorithm is defined and explained using an illustrative example in Section \ref{ACHCI}.
Section \ref{benchmark} describes competing benchmark algorithms, which include the NN and NI heuristics, as well as population-based metaheuristic approaches that have been successfully implemented on microcontroller-based systems, namely the Genetic Algorithm (GA) \cite{griffiths1998microcontroller,wu2024decentralized} and Ant Colony Optimization (ACO) \cite{scheuermann2004fpga,dahmane2020ant}.
After this, section \ref{computational} compares their performance with that of ACHCI using 57 modified TSPLIB instances that represent a broad range of real-world NETSPs \cite{reinhelt2014tsplib}.
% After defining the proposed ACHCI algorithm and its competing benchmark algorithms, this paper evaluates their performance on 57 modified TSPLIB instances that represent a broad range of real-world NETSPs \cite{reinhelt2014tsplib}. 
These TSPLIB instances are modified by either using the $\mathcal{L}_1$ norm as the cost function to capture urban layouts with grid-based road infrastructure, or by inserting impassable separators that simulate obstacles that create varying levels of distortion in the Euclidean space.
Additionally, in Section~\ref{caseStudy}, two warehouse scenarios from the MAPF benchmark are used to illustrate the effectiveness of ACHCI in addressing NETSPs encountered in real-world logistics settings.

Through extensive computational experiments, this paper shows that the ACHCI heuristic consistently outperforms the NN and NI heuristics across the modified test sets, producing lower-cost tours in 88\% and 99\% of the test cases, with average cost reductions of 11\% and 9\%, respectively.
ACHCI also achieves lower tour costs than GA and ACO in 87\% and 94\% of instances within a 60-second computation time budget, and on average, produces tours that were 8.28 times lower than GA and 27\% lower than ACO tours.

\textcolor{black}{
For completeness, the appendix tabulates ACHCI solution costs for multiple TSPLIB instances with varying causes of non-Euclidean space, comparing its performance with NN, NI, GA and ACO.
Further, the \textcolor{black}{performance of ACHCI is compared} with that of the LKH heuristic in the appendix \textcolor{black}{using a matched wall-clock time budget to assess} the effectiveness of the lightweight heuristic with the near-optimal solutions generated by LKH.
The appendix also examines alternatives to the MDS for embedding the non-Euclidean points in a Euclidean space and the effect of using the insertion cost ratio when selecting points for insertion instead of incremental insertion costs.
}

The results \textcolor{black}{justify the algorithmic design choices and show that the ACHCI algorithm outperforms competing lightweight heuristics, suggesting that ACHCI is a promising approach for onboard tour computation on resource-limited autonomous agents.
While LKH generally achieves lower-cost tours, ACHCI provides deterministic real-time tour construction through a comparatively simple algorithmic procedure compatible with embedded onboard deployment constraints.
This directly translates to enhanced decentralized control for faster task completion, lower real-world costs, shorter travel distances and lower energy consumption, indirectly reducing charging-related downtime when applicable.
}

\section{Non-Euclidean TSP}\label{NETSP}

Consider $n$ locations that must be sequentially visited by an agent such that each location is only visited once, and the agent returns to the first visited location at the end of the tour.
In the illustrative example of a warehouse shown in Fig. \ref{fig:warehouse}, every point marked by red dots on the warehouse floor must be visited by the robot shown near the bottom right.
Since inventory shelves prevent straight-line travel, non-Euclidean costs exist between visit locations, and therefore the Euclidean CHCI algorithm cannot be directly used. 
The proposed ACHCI algorithm addresses this by first finding a representative set of points in a Euclidean space, with pairwise distances that approximate these non-Euclidean costs. 
The convex hull of these points initializes the sub-tour shown in Fig. \ref{fig:initialTour}, after which the non sub-tour points are added in increasing order of their true non-Euclidean insertion cost ratios, producing the tour shown in Fig. \ref{fig:TSPNEsolution}.

Formally, the TSP is defined over a fully connected graph $\mathcal{G} := (\mathcal{V}, A)$, where $\mathcal{V} := \{v_1,v_2,\dots,v_n\}$ denotes the set of locations or nodes to be visited exactly once by the agent.
The directed arc set $A := \{(i,j) \in \mathcal{V} \times \mathcal{V} \mid i \neq j\}$ is the set of edges that connect every ordered pair of nodes in $\mathcal{V}$.

The decision to travel directly from node $i$ to node $j$ is denoted by setting the value of binary decision variable $x_{ij}$ to 1, with 0 being the complementary case. 
The cost function $c : A \rightarrow \mathbb{R}^+$ defines the cost to be minimized, and the cost associated with choosing edge $(v_i,v_j)$ is given by $c_{ij}$.
The objective function defined by Eq.~\eqref{eq:classic_obj} minimizes the total travel cost incurred over all selected edges:
\begin{subequations}  \label{eq:ClassicTSP}
\allowdisplaybreaks
\begin{align}
    \label{eq:classic_obj}     
    &\min_{x_{ij}} \sum_{i \in \mathcal{V}} \sum_{\substack{j \in \mathcal{V} \\ j \ne i}} c_{ij} x_{ij} \\
    \label{eq:classic_binary}
    &x_{ij} \in \{0,1\} \quad \forall i,j \in \mathcal{V}, ~i \ne j \\
    \label{eq:visit_once_out}
    &\sum_{\substack{j \in \mathcal{V} \\ j \ne i}} x_{ij} = 1 \quad \forall i \in \mathcal{V} \\
    \label{eq:visit_once_in}
    &\sum_{\substack{j \in \mathcal{V} \\ j \ne i}} x_{ji} = 1 \quad \forall i \in \mathcal{V} \\
    \label{eq:mtz_order}
    &u_i - u_j + n x_{ij} \leq n - 1 \quad \forall i \ne j,~ i,j \in \mathcal{V} \setminus \{v_1\} \\
    \label{eq:mtz_bounds}
    &1 \leq u_i \leq n - 1 \quad \forall i \in \mathcal{V} \setminus \{v_1\}
\end{align}
\end{subequations}

% Constraint~\eqref{eq:classic_binary} enforces the binary nature of the decision variables, ensuring that each edge is either selected or not selected in the TSP tour. 
Constraints~\eqref{eq:visit_once_out} and \eqref{eq:visit_once_in} guarantee that each node is visited exactly once by requiring exactly one utilized outgoing and one incoming edge per node. 

To enforce tour connectivity by eliminating multiple sub-tours in the resulting TSP tour, the Miller–Tucker–Zemlin formulation introduces auxiliary variables $u_i$ for all $i \in \mathcal{V} \setminus \{v_1\}$ \cite{miller1960integer}.
These sub-tour elimination constraints in~\eqref{eq:mtz_order} enforce a valid sequencing of visits, while bounds in~\eqref{eq:mtz_bounds} restrict the range of these variables to ensure feasible ordering. 

When the cost function $c$ defines a non-Euclidean cost for the relation between every pair of nodes, Eq. \eqref{eq:ClassicTSP} is called the non-Euclidean TSP.
Let the non-Euclidean arc costs be captured by a cost matrix $C\in\mathbb{R}^{n\times n}$ whose entries are defined as $C_{ij} = c(v_i,v_j) ~ \forall (v_i,v_j) \in A$.

\textcolor{black}{
The vast majority of NETSPs that arise from warehouse navigation or obstacle-laden environments define costs as shortest-path distances over an underlying graph with nonnegative edge costs.
In such cases, the cost matrix satisfies the triangle inequality, $C_{ij} \leq C_{ik} + C_{kj}$ for all $i,j,k \in \mathcal{V}$.
Further, when the underlying environment is undirected, i.e., the direction of travel does not affect cost, these shortest-path costs satisfy $C_{ij}=C_{ji}$ for all $i,j\in\mathcal{V}$, resulting in a symmetric cost matrix $C$.
When these conditions are met, the cost matrix is said to be metric.
}

\textcolor{black}{
Asymmetric cost matrices arise in directed arc sets, when travel costs depend on direction, such as in the presence of asymmetric turn penalties, one-way road segments or aisle systems, or other direction dependent constraints \cite{goutham2023optimal}. 
In such cases, the Jonker--Volgenant transformation \cite{jonker1983transforming} may be applied to convert an asymmetric cost matrix $C\in\mathbb{R}^{n\times n}$ into an equivalent symmetric matrix $\tilde{C}\in\mathbb{R}^{2n\times 2n}$, after which the proposed ACHCI algorithm described in the next section can be applied to the transformed cost matrix.
}

\section{Adapted Convex Hull Cheapest Insertion Algorithm}\label{ACHCI}

The ACHCI heuristic first uses the metric MDS, applying principal component analysis (PCA) to find a set of points in 2D space whose pairwise Euclidean distance approximates the non-Euclidean arc cost function \cite{torgerson1952multidimensional,anderson2003canonical}.
This is initiated by assigning one of the TSP points to be the origin in a new Euclidean space without loss of generality, and then finding the coordinates of the remaining $n-1$ points such that their pairwise Euclidean distances are exactly their non-Euclidean costs.

Assuming a full rank \textcolor{black}{metric} cost matrix $C$, these points are first obtained in an $(n-1)$-dimensional Euclidean space, after which they are projected to a 2D space.
Let the Euclidean coordinate equivalent of the node $v_i \in \mathcal{V}$ be represented by $x_i \in \mathbb{R}^{n-1}$.
In this paper, the notation $x_i \leftrightarrow v_i$ is used to indicate this mapping.
For some $\{x_j,x_k\} \in \mathbb{R}^{n-1}$, the relative position vectors of $x_i$ and $x_j$ with respect to $x_k$ are $(x_i-x_k)$ and $(x_j - x_k)$ respectively.
If the Euclidean distance between these points is identical to their respective non-Euclidean costs as defined by cost matrix $C$, it can be verified that their inner product satisfies
\begin{equation}\label{eq:dotProduct}
        \langle x_i - x_k , x_j-x_k\rangle = (C_{ik}^2+C_{kj}^2-C_{ij}^2)/2
\end{equation}

This relation between the desired Euclidean coordinates and the cost matrix $C$ is leveraged to calculate the coordinates of each point relative to the origin in the $n-1$ dimensional Euclidean space.
Without loss of generality, pick position vector $x_1 \leftrightarrow v_1$ as the origin and define $\Bar{x}_i:=x_i-x_1$ as the relative position vector of $x_i$ with respect to $x_1$.
Let the ordered collection of all relative position vectors form a column matrix $\Bar{X} \in \mathbb{R}^{(n-1)\times(n-1)}:= [\Bar{x_2},\Bar{x_3},...,\Bar{x_n}]$.
% , with $\Bar{x}_1=\Bar{0}$.

Define the Gramian matrix $M\in \mathbb{R}^{(n-1)\times(n-1)}$ by $M_{ij}:=\langle \Bar{x_i}, \Bar{x_j}\rangle$. 
Then by definition,
 \begin{align}
    \label{eq:Gram}
    M &= \bar{X}^\top \bar{X} 
\end{align}

\begin{figure}[b]
    \centering
  \subfloat[Non-Euclidean point-cloud $V$ with 4 impassable separators \label{Im1a: PointCloud}]{%
       \includegraphics[trim =5mm 5mm 0mm 0mm, clip, width=0.7\linewidth]{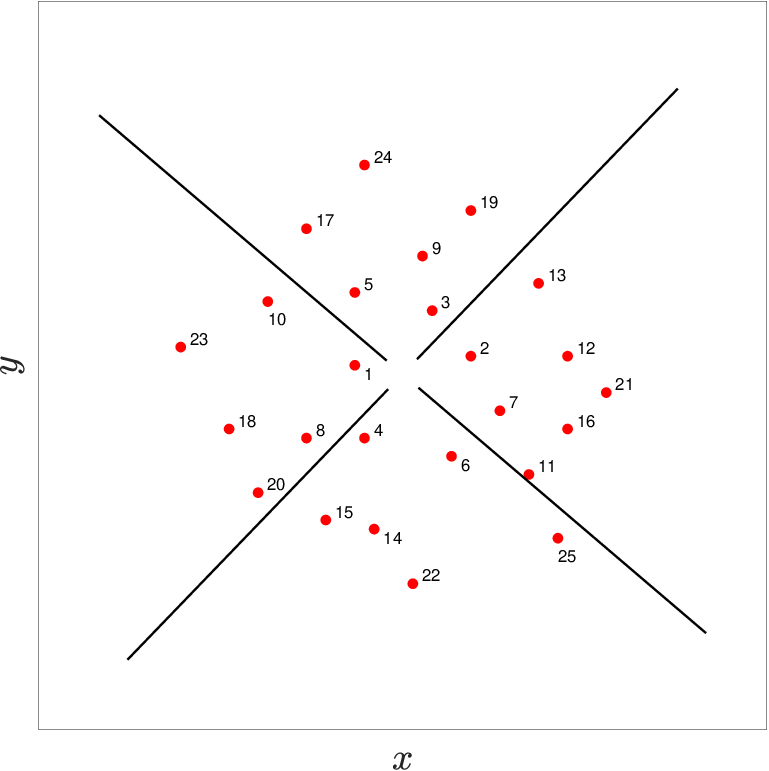}}
       \\
       \vspace{5mm}
  \subfloat[Convex hull of Euclidean 2D approximate coordinates $\widetilde{X}$\label{Im1b: MDSAppx}]{%
        \includegraphics[trim =5mm 5mm 0mm 0mm, clip, width=0.7\linewidth]{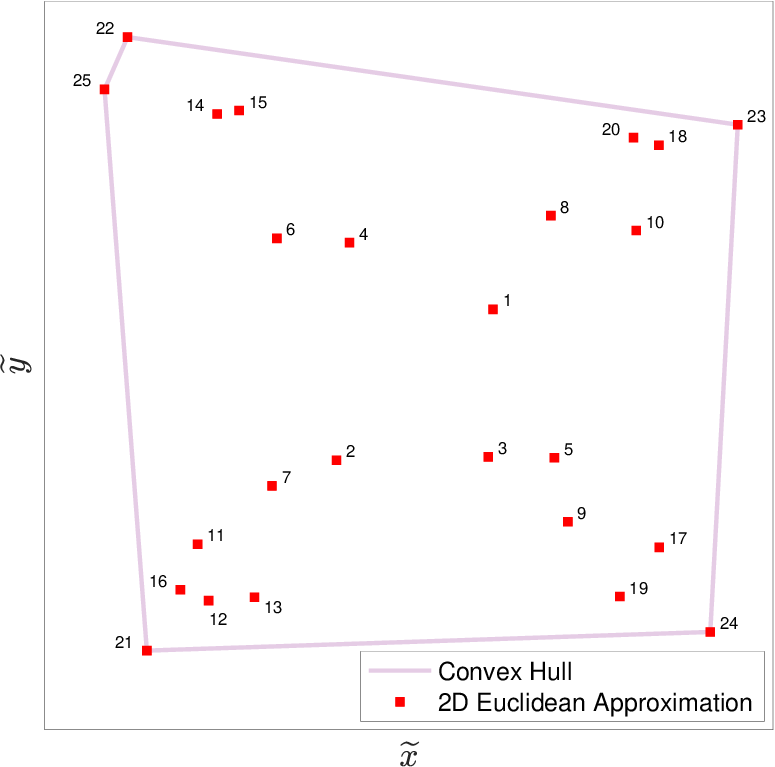}}
  \caption{Using multi-dimensional scaling (MDS) to find the Euclidean 2D approximation of a non-Euclidean instance}
  \label{fig1: MDS} 
  \vspace{-2mm}
\end{figure}

Each entry of  matrix $M$ can be calculated using the cost matrix $C$ and Eq. (\ref{eq:dotProduct}).
The eigenvalue decomposition of $M$ results in eigenvector matrix $Q\in\mathbb{R}^{(n-1)\times (n-1)}$ and diagonal eigenvalue matrix $\Lambda \in \mathbb{R}^{(n-1)\times (n-1)}$ as shown in  Eq. (\ref{eq:eigValDecomp}).
Because Gram matrices are positive semi-definite, the eigenvalues of $M$ are necessarily non-negative.
If $\Sigma$ is the diagonal singular value matrix so that $\Lambda = \Sigma^\top \Sigma$, then,
% Because $\Sigma$ is a diagonal matrix, Eq. (\ref{eq:eigValDecomp}) is manipulated to obtain $M = (Q\Sigma)^\top(Q\Sigma)$ in  Eq. (\ref{eq:eigValDecomp}).
 \begin{align}
    \label{eq:eigValDecomp}
    M = Q\Lambda Q^\top = Q\Sigma^\top \Sigma Q^\top &= (\Sigma Q^\top)^\top(\Sigma Q^\top)
\end{align}
Using Eq. (\ref{eq:Gram}) and (\ref{eq:eigValDecomp}), it is clear that $\bar{X} = \Sigma Q^\top$ defines a set of point coordinates that have Euclidean distances that are identical to the defined non-Euclidean costs.
However, they are in an $n-1$ dimensional space, where, for $n>2$, obtaining the convex hull initiated TSP sub-tour is not possible.
For this reason, PCA is applied to obtain the set of two-dimensional point coordinates with pairwise distances that best approximate the non-Euclidean cost function.
The two largest eigenvalues $\lambda_{max_1}, \lambda_{max_2}$ are chosen to form $\widetilde{\Sigma}\in\mathbb{R}^{2\times 2} := $ diag$(\sqrt{\lambda_{max_1}}, \sqrt{\lambda_{max_2}})$.
Let $\widetilde{Q}\in \mathbb{R}^{(n-1)\times 2}$ contain their respective eigenvectors.

The approximated 2D coordinates are then obtained as the columns of $\widetilde{X}\in\mathbb{R}^{2\times(n-1)}$:
\begin{equation*}
        \widetilde{X} = \widetilde{\Sigma}\widetilde{Q}^T
\end{equation*}
This procedure is illustrated for a set of 25 enumerated points that are separated by impassable line segments, creating a non-Euclidean point cloud as shown in Fig. \ref{Im1a: PointCloud}. 
The 2D Euclidean approximate coordinates that define matrix $\widetilde{X}$ are shown in Fig. \ref{Im1b: MDSAppx} where the pairwise Euclidean distances approximate the pairwise shortest distances of the original non-Euclidean point cloud.
Also shown is the convex hull of these 2D points that initiates the sub-tour of the ACHCI heuristic.
\begin{figure}[t]
    \centering
  \subfloat[Initialized sub-tour $T_0$ as the convex hull nodes of $\widetilde{X}$ \label{Im1c: CHullInit }]{%
        \includegraphics[trim =5mm 5mm 0mm 0mm, clip, width=0.7\linewidth]{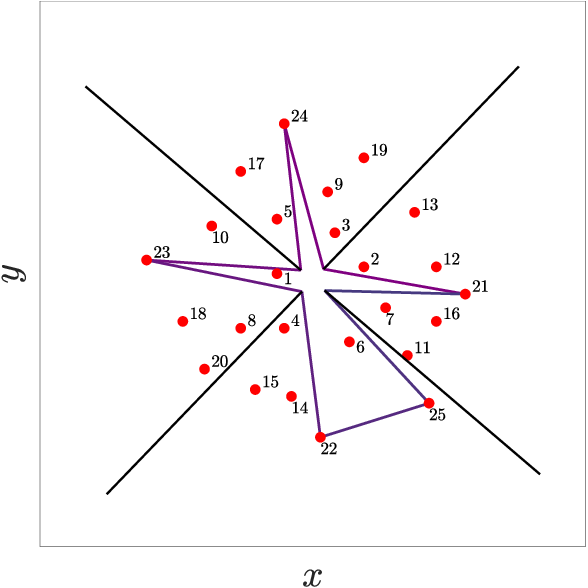}}
       \\
       \vspace{5mm}
  \subfloat[Completed non-Euclidean tour \label{Im1d: ACHCIsol}]{%
        \includegraphics[trim =5mm 5mm 0mm 0mm, clip, width=0.7\linewidth]{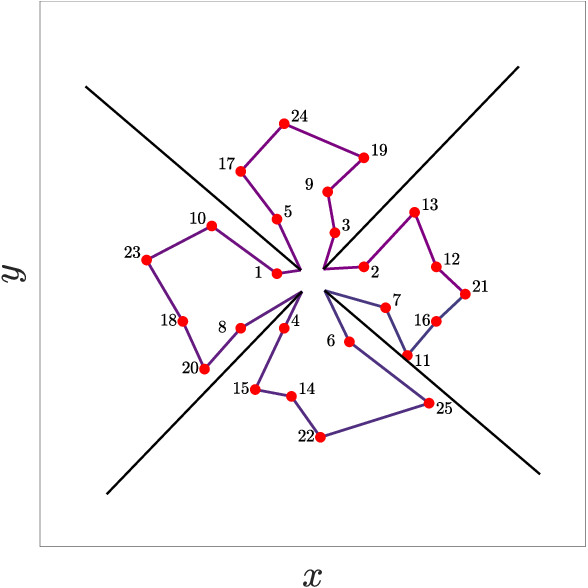}}
  \caption{ACHCI sub-tour initiation and complete tour for a non-Euclidean TSP instance with 25 points}
  \label{fig2: neTSP CHULL} 
\end{figure}

\newpage
The ACHCI algorithm is summarized as:
\begin{enumerate}[\textit{Step} 1:]
\item Use MDS to obtain 2D Euclidean approximate coordinates that define matrix $\widetilde{X}$
\item Initiate sub-tour as the convex hull nodes of $\widetilde{X}$. Next, discard the Euclidean point approximation and use the true non-Euclidean arc costs of the $C$ matrix for the remaining steps of the ACHCI algorithm.
\item Find consecutive nodes $v_i,v_j \in V$ in the sub-tour and $v_k\in V$ not in the sub-tour, that minimizes non-Euclidean insertion cost ratio $(C_{ik} + C_{kj})/C_{ij}$.
    % For each k, find (i,j) such that $c_{ik} + c_{kj} - c_{ij}$ is minimal. 
    % From each (i,k,j) triplet, find the one such that $c_{ik} + c_{kj}/c_{ij}$ is minimal. 
\item Insert $v_k$ between $v_i$ and $v_j$, updating the sub-tour.
\item Repeat \textit{Steps} 3 and 4, until all nodes are in the tour, thus obtaining the NETSP solution.
\end{enumerate}

For the example set of 25 non-Euclidean points, the sub-tour is initiated using the convex hull boundary points of $\widetilde{X}$ as shown in Fig. \ref{Im1c: CHullInit }, after which the Euclidean coordinate approximation is discarded.
The insertion cost ratios of \textit{Step 3, 4} and \textit{5} use the true non-Euclidean cost matrix $C$ to obtain the complete tour, which is shown in Fig. \ref{Im1d: ACHCIsol}.

\textcolor{black}{
Apart from the implemented MDS method, other methods exist for embedding a non-Euclidean cost matrix into a 2D Euclidean space.
Among these, Uniform Manifold Approximation and Projection (UMAP)\cite{mcinnes2018umap, healy2024uniform} and t-distributed Stochastic Neighbor Embedding (t-SNE) \cite{van2008visualizing,vandermaaten2014accelerating} are dimensionality reduction techniques that have been successfully applied to high-dimensional distance matrices, prioritizing the preservation of local neighborhood structure \cite{mcinnes2018umap, healy2024uniform}.
However, ACHCI specifically requires an embedding that preserves global distance structure, since the convex hull of the embedded points is used to initialize the sub-tour. 
For this reason, MDS is chosen because it explicitly minimizes global distance distortion, making it more appropriate for identifying boundary-like points.
In contrast, UMAP and t-SNE do not guarantee preservation of global pairwise distances, and consequently, a point that is geometrically central with respect to the non-Euclidean cost structure could appear on the convex hull boundary as an artifact of its local cluster arrangement, potentially degrading the ACHCI initialization.
}

\textcolor{black}{
Alternative formulations also exist for the criterion used in \textit{Step 3} of the ACHCI algorithm when selecting which unvisited node should be inserted into the sub-tour.
The incremental insertion cost criterion $C_{ik} + C_{kj}-C_{ij}$ selects the node and insertion position that minimizes the overall increase in tour length. 
In contrast, ACHCI employs the insertion cost ratio $(C_{ik} + C_{kj})/C_{ij}$, which penalizes insertions that cause large relative detours with respect to the insertion edge \cite{rosenkrantz1977analysis}.
This ratio criterion has been shown to empirically yield favorable tour costs for Euclidean TSPs \cite{huang2017investigating}.
}

\textcolor{black}{
For completeness, Appendix A provides a comparison of ACHCI performance under the alternative embeddings of UMAP and t-SNE along with a comparison of the two insertion criteria discussed.
It is shown that the selection of MDS with cost ratio based insertion empirically yields lower tour costs for the vast majority of NETSPs being studied.
}

\section{Benchmark algorithms}\label{benchmark}

Two heuristic algorithms and two metaheuristic algorithms are selected as benchmarks to evaluate the performance of the proposed ACHCI algorithm. 
The Nearest Neighbor (NN) and Nearest Insertion (NI) algorithms are greedy heuristics that rely on simple selection rules and are commonly used in constrained TSP applications due to their speed and ease of implementation \cite{grigoryev2018solving, bai2020efficient, kumar2013traveling}. 
The Genetic Algorithm (GA) and Ant Colony Optimization (ACO) are population-based metaheuristics that explore the solution space more broadly by iteratively updating a set of candidate solutions \cite{halim2019combinatorial}.
Although GA and ACO are more computationally demanding than NN and NI, they are also included as benchmark algorithms because they have been successfully implemented on microcontroller-based systems, demonstrating their feasibility for onboard deployment in autonomous agents tasked with route planning \cite{griffiths1998microcontroller, wu2024decentralized, scheuermann2004fpga, dahmane2020ant}.
\textcolor{black}{
As detailed in Section \ref{sec:intro}, LKH and Christofides algorithms are not considered primary onboard benchmark algorithms since there are no reported deployments onboard microcontroller or embedded systems. 
Nevertheless, Appendix B compares the effectiveness of the ACHCI heuristic against LKH.
}

\subsection{Nearest Neighbor Heuristic}
Starting from a predefined or randomly selected initial node, the NN algorithm assigns the unvisited node associated with the lowest cost as the next node.
This is repeated until all nodes are contained in the tour after which the starting location is visited again.
The NN heuristic is as described below:
\begin{enumerate}[\textit{Step} 1:]
    \item Initiate the sub-tour with the point closest to the centroid as the starting location
    \item Append the sub-tour with the location with lowest non-Euclidean cost from the location that is at the end of the current sub-tour
    \item Repeat step 2 until all locations have been included
    \item Return to the starting location
\end{enumerate}

\subsection{Nearest Insertion Heuristic}

In contrast with the NN heuristic that is initiated with a single location, the NI heuristic starts with a sub-tour of 2 nodes that are identical, representing the start and end of the TSP tour. It then expands it by inserting unvisited nodes in increasing order of insertion cost:
\begin{enumerate}[\textit{Step} 1:]
\item Initialize a sub-tour as a starting location to itself.
\item Find consecutive nodes $(v_i,v_j)$ in the sub-tour, and $v_k$ not in the sub-tour, that minimizes non-Euclidean \textcolor{black}{incremental} insertion cost $C_{ik} + C_{kj}-C_{ij}$.
\item Insert $v_k$ between $v_i$ and $v_j$, updating the sub-tour.
\item Repeat \textit{Step} 2 and \textit{Step} 3, to obtain a TSP solution.
\end{enumerate}

\subsection{Genetic Algorithm}

The GA is a popular metaheuristic algorithm, inspired by the mechanisms of natural selection and genetics \cite{larranaga1999genetic}.
The GA process is initialized with a \textit{population} of $m$ randomly generated tours, known as chromosomes.
The quality of each chromosome is defined by its \textit{fitness function}, assigned as the inverse of the TSP tour length.
In subsequent iterations, these chromosomes are modified to find solutions with higher fitness value.
At any iteration, the incumbent solution is defined as the chromosome with the highest fitness value or lowest tour cost found up to that iteration.
To ensure that the incumbent solution does not degrade as the iterations proceed, a defined number of high fitness chromosomes, called the \textit{elite population} are carried over to the next generation without modification.

In each iteration, new tours called \textit{offspring} are created by modifying some selected \textit{parent} chromosomes of the current population.
To ensure that promising parents have a higher chance of being selected and contributing to the next population, each chromosome $j$ with a fitness value $f_j$ is assigned a fitness proportionate probability given by $P_j = f_j/ \sum_{i = 1}^m f_i$.
These probabilities are used by a \textit{Roulette Wheel Selection} method to sample high fitness parents for the next generation \cite{holland1992adaptation}.

The elite population and the Roulette Wheel selected chromosomes form the $m$ parents for the next generation.
Of these parents, a fraction $p_1$ is randomly selected for the \textit{Ordered Crossover} operation, where segments of two parent chromosomes are exchanged to produce two offspring.
To ensure that duplicate points do not feature in the resulting offspring, each offspring is constructed by choosing a sub-sequence of one parent while preserving the relative order of remaining points of the other \cite{davis1985applying}.
Next, a fraction $p_2$ of the resulting population is randomly sampled for the process of \textit{Swap Mutation} whereby two random locations within a chromosome are selected and swapped to form a new offspring \cite{louis2000case}.
Of the resulting population, a fraction $p_3$ is then sampled for the \textit{Inversion Mutation}, where a sub-tour segment of a parent chromosome is randomly chosen and its order is reversed \cite{holland1992adaptation}.

For a predefined time limit, the operations of selection, crossover, and mutation iteratively explore the search space of TSP solutions to improve the incumbent solution.
Upon termination, the incumbent solution is the output tour.

\subsection{Ant Colony Optimization}
The ACO is another nature-inspired metaheuristic algorithm that mimics how a colony of ants finds the shortest paths to food sources based on distance and the strength of pheromone trails laid by individual ants \cite{dorigo1996ant}.
While the GA uses stochastic processes to vary tour costs without explicitly evaluating edges in the arc set $A$, the ACO assigns pheromone strengths and heuristic values to the edges to inform exploration.

The ACO is initialized by computing local heuristic values $\eta_{ij}$ for each arc $(v_i,v_j) \in A$, assigned as the inverse of the cost $c(v_i,v_j)$ for $v_i\neq v_j$, and zero otherwise.
The pheromone value $\tau_{ij}~\forall~(v_i,v_j) \in A$ is initially assigned a negligible value called the primary tracing, and in subsequent iterations, $\tau_{ij}$ will be updated to reflect the effect of edge $(v_i,v_j)$ on tour costs.

In each iteration, ants start from the randomly chosen location $v_i$ and then build tours by sampling unvisited locations $v_j$, based on the transition probability $P_{ij}$ of arcs $(v_i,v_j)\in A$:
\begin{equation}\label{eq:probabilisticEdge}
    P_{ij} = \frac{\tau_{ij}^\alpha \eta_{ij}^\beta}{\sum_{k\notin sub-tour}\tau_{ik}^\alpha \eta_{ik}^\beta}
\end{equation}
The influence of pheromone and heuristic information on tour edge selection is controlled by exponential parameters $\alpha$ and $\beta$ respectively \cite{dorigo1999ant}.
Let $c_a$ represent the cost of the tour sampled by ant $a$ of colony $\mathcal{A}$, obtained by sampling locations based on Eq.(\ref{eq:probabilisticEdge}).
After tours have been constructed for the ant colony, pheromone levels $\tau_{ij}$ on each edge $(v_i,v_j)$ are updated before the next iteration:
\begin{equation}\label{eq:pheromone}
    \tau_{ij}  \gets (1-\rho)\tau_{ij} + \sum_{a\in\mathcal{A}}c_a^{-1}
\end{equation}
The evaporation rate parameter $\rho \in (0,1)$ discounts $\tau_{ij}$ in the next iteration to encourage exploration.

\sloppy
The iterative update of transition probabilities and pheromone trails results in an increased exploration of tours that exploit edges associated with lower tour costs.
This guides the algorithm in finding improved tours based on the likelihood of an edge constituting the optimal tour, while the evaporation of pheromones encourages exploration of the entire search space to prevent premature convergence.
The algorithm is terminated based on a convergence criteria such as the computation time limit, and the incumbent tour is provided as the output.

\section{Computational Experiments}\label{computational}
\begin{figure}[b]
\centering
  \subfloat[Manhattan or $\mathcal{L}_1$ norm\label{t70_L1}]{%
       \includegraphics[trim =6mm 6mm 0mm 0mm, clip, width=0.45\linewidth]{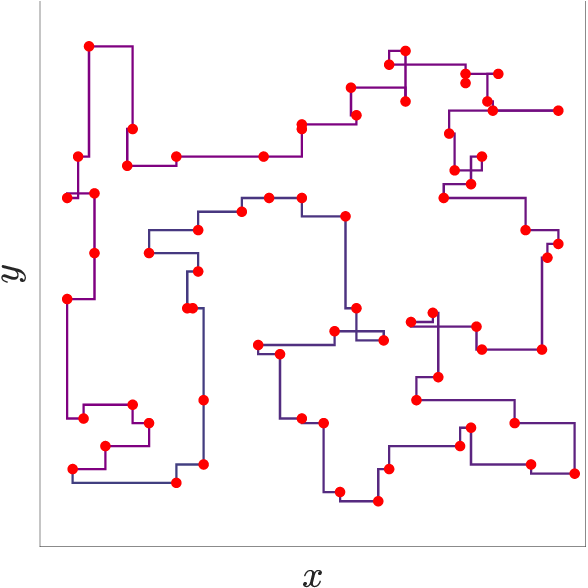}}
    \hspace{.02\linewidth}
  \subfloat[4 separators\label{Im 4a: NETSP}]{%
       \includegraphics[trim =6mm 6mm 0mm 0mm, clip, width=0.45\linewidth]{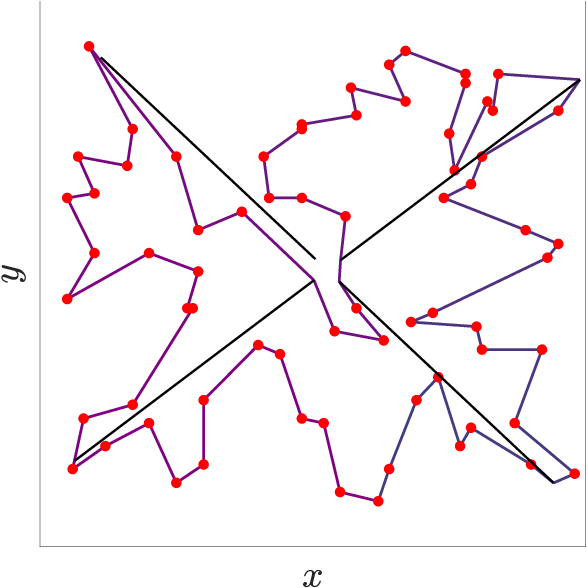}}
    \hspace{.02\linewidth}
  \caption{Two types of non-Euclidean modifications on the TSPLIB instance \textit{`t70'}, shown with their ACHCI tours.}
  \label{Im:Example ACHCI} 
\end{figure}
\subsection{Non-Euclidean Test Cases}
To compare the effectiveness of the ACHCI algorithms with the 4 benchmark algorithms, sufficiently diverse test instances were not readily available for the NETSP.
For this reason, the popular TSPLIB instances \cite{reinhelt2014tsplib} are modified to create non-Euclidean point clouds.
In the first type of non-Euclidean modification, the $\mathcal{L}_1$ or Manhattan norm is defined as the cost between two locations in a TSPLIB point cloud \cite{petrovic2014geometry}. %Energy-efficient automated material handling systems
In the second case, impassable separators are added as environmental constraints.

The following steps ensure the reproducibility of adding separators to the TSPLIB point cloud:
\begin{enumerate}[\textit{Step} 1:]
    \item Load a TSPLIB point cloud with 2D Cartesian coordinates.
    \item Find the centroid of the point cloud.
    \item Sort the points in order of increasing distance from the centroid.
    \item Draw a line segment from the centroid to the farthest point. Trim its length by 5\% from both ends. This line segment functions as the impassable separator.
    \item For $k$ equiangular separators, rotate copies of the separator obtained in \textit{Step 4} about the centroid by multiples of $2\pi/k$ radians. 
\end{enumerate}

The ACHCI solution to the \textit{`t70'} instance outfitted with the $\mathcal{L}_1$ norm is shown in Fig. \ref{t70_L1}, where it can be seen that paths are only permissible along the $x$ and $y$ directions.
A deviation from the Euclidean straight-line path is also seen in the case with 4 added separators, as shown in Fig. \ref{Im 4a: NETSP}.
The distortion of the Euclidean space caused by non-Euclidean cost functions is quantified in the literature \cite{boyaci2021vehicle,cole1968quantitative} by the deviation factor (DF):
\begin{equation}\label{Eq: DF}
    DF = \binom{|V|}{2}^{-1} \sum_{\substack{(v_i,v_j)\in A\\ v_i \neq v_j}} \frac{\Delta(v_i,v_j)}{\delta(v_i,v_j)}
\end{equation}
In Eq. (\ref{Eq: DF}), for each pair of points $(v_i,v_j)\in A$, the Euclidean distance is denoted by $\delta(v_i,v_j)$ while $\Delta(v_i,v_j)$ is the true non-Euclidean cost.
If this cost is defined by the $\mathcal{L}_1$ norm, then $\Delta(v_i,v_j)=|x_{i_1}-x_{j_1}|+|x_{i_2}-x_{j_2}|$, where $(x_{i_1},x_{i_2})$ and $(x_{j_1},x_{j_2})$ define the coordinates of $v_i$ and $v_j$ respectively.
When instead, obstacles like impassable separators cause the deviation from the straight-line path, the shortest true path length between pairs of points in $(v_i,v_j)\in A$ is computed using Dijkstra's shortest path algorithm \cite{dijkstra2022note} to find $\Delta(v_i,v_j)$.
 
\begin{table}[b]
\centering
\caption{Metaheuristic hyperparameter settings}\label{HyperParameterSettings}
    \resizebox{0.8\columnwidth}{!}{
    \begin{tabular}{ll}
\textbf{Genetic Algorithm}                                     &                       
    \\ \hline
\multicolumn{1}{|r|}{Population Size:} 
    & \multicolumn{1}{l|}{1000} 
    \\ \hline
\multicolumn{1}{|r|}{Elite Count:}     
    & \multicolumn{1}{l|}{10} 
    \\ \hline
\multicolumn{1}{|r|}{Ordered Crossover Fraction $p_1$:}    
    & \multicolumn{1}{l|}{0.9} 
    \\ \hline
\multicolumn{1}{|r|}{Swap Mutation Fraction $p_2$:}     
    & \multicolumn{1}{l|}{0.02} 
    \\ \hline
\multicolumn{1}{|r|}{Inversion Fraction $p_3$:}     
    & \multicolumn{1}{l|}{0.07} 
    \\ \hline
\textbf{\rule{0pt}{4ex} Ant Colony Optimization}                                   &                       
    \\ \hline
\multicolumn{1}{|r|}{Colony Size}                 
    & \multicolumn{1}{l|}{1000} 
    \\ \hline
\multicolumn{1}{|r|}{Primary tracing:}                 
    & \multicolumn{1}{l|}{$10^{-4}$} 
    \\ \hline
\multicolumn{1}{|r|}{Exponential Pheromone Parameter $\beta$:}                 
    & \multicolumn{1}{l|}{3} 
    \\ \hline
\multicolumn{1}{|r|}{Exponential Heuristic Parameter $\alpha$:}                 
    & \multicolumn{1}{l|}{5} 
    \\ \hline
\multicolumn{1}{|r|}{Evaporation Coefficient $\rho$:}                 
    & \multicolumn{1}{l|}{0.15} 
    \\ \hline
    \end{tabular}}
\end{table}

\subsection{Performance Analysis}

To analyze the performance of the ACHCI heuristic, extensive computational experiments were conducted in a Matlab R2022a environment on an Intel Xeon E5-2680 v4 CPU clocked at 2.4 GHz at the Ohio Super Computer \cite{OhioSupercomputerCenter1987}.
The utility of the approximate algorithms compared in this paper lies in their ability to obtain solutions quickly, and for this reason, each algorithm is executed up to a time limit of 60 seconds.
The hyperparameters of the metaheuristic algorithms are listed in Table \ref{HyperParameterSettings}, and influence the convergence of incumbent solutions.
To account for the stochasticity involved in the GA and ACO algorithms, each algorithm is repeated 28 times for each TSPLIB case and the distribution of solutions is noted upon termination at 60 seconds.

\begin{figure}[t]
    \centering
    % \hspace{20px}
  \subfloat[TSPLIB instance \textit{`roC100'} with $\mathcal{L}_1$ norm\label{Im:roC100convergence}]{%
        \includegraphics[trim =40mm 100mm 40mm 100mm, clip, width=0.9\linewidth]{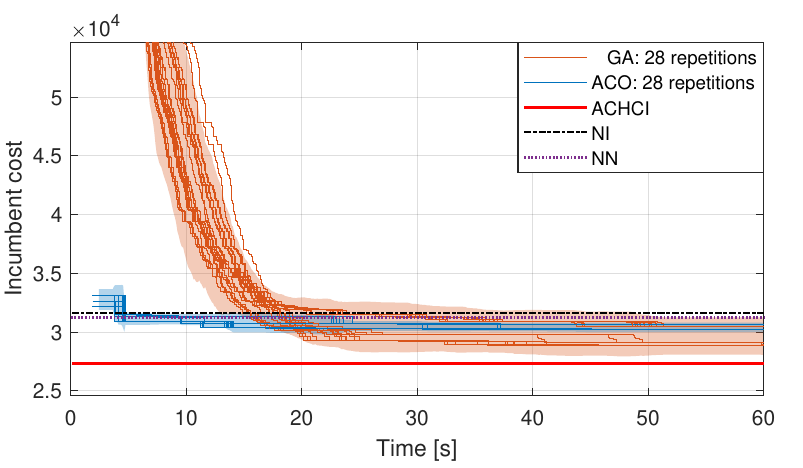}}\\
        \vspace{3mm}
        % \hspace{0px}
  \subfloat[TSPLIB instance \textit{`roB150'} with 4 separators\label{Im:roB150convergence}]{%
        \includegraphics[trim =40mm 100mm 40mm 100mm, clip, width=0.9\linewidth]{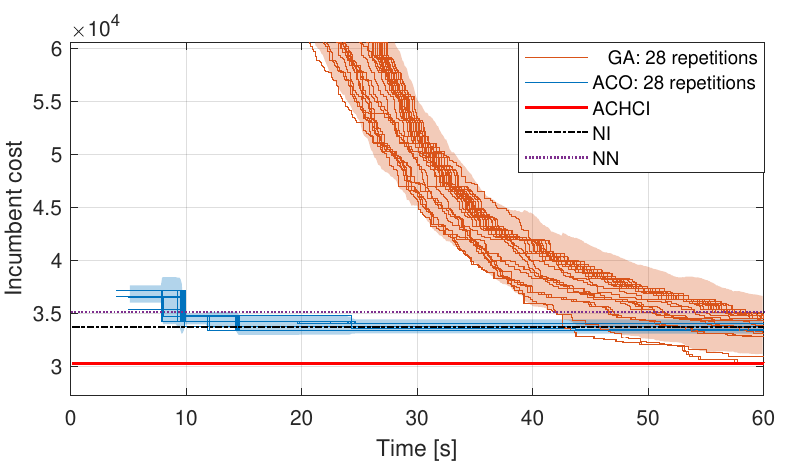}}
  \caption{Convergence of metaheuristic algorithms}
  \label{fig3}
  \vspace{-7mm}
\end{figure}

The convergence of GA and ACO incumbent solutions along with their 95\% confidence interval is shown in Fig. \ref{Im:roC100convergence} for the case where the $\mathcal{L}_1$ norm is applied to the TSPLIB instance \textit{`roC100'} with 100 points.
The GA only uses tour cost as a metric for tour improvement which results in rapid rates of convergence.
In contrast, ACO samples and updates pheromone values $\tau_{ij}$ using Eqs. (\ref{eq:probabilisticEdge}) and (\ref{eq:pheromone}), resulting in higher computational demand per iteration.
While this results in a slower convergence rate, its incumbent solution costs are initially lower than that of GA.
In Fig. \ref{Im:roB150convergence} where 4 separators were added to the 150 points of instance \textit{`roB150'}, it is seen that despite the larger number of iterations, the initial performance of the ACO is still superior.
% While the metaheuristic approaches iteratively improve a population of solutions, heuristic approaches produce only one solution for each instance.

Performance comparisons are conducted for 57 TSPLIB instances, with the number of points varying from 51 to 1,400.
For each TSPLIB instance, one non-Euclidean case is created by assigning costs using the $\mathcal{L}_1$ norm, and 3 cases are created by adding 4, 16 or 64 impassable separators.
The resulting variation in DFs and number of points is used to study of the effectiveness of the ACHCI heuristic, with performance comparisons tabulated in Table 2.

\begin{figure}[b]
    \centering
    % \hspace{20px}
  \subfloat[Cost ratios for various DF\label{H_ratio_DF}]{%
        \includegraphics[trim =69mm 103mm 75mm 107mm, clip, width=0.49\linewidth]{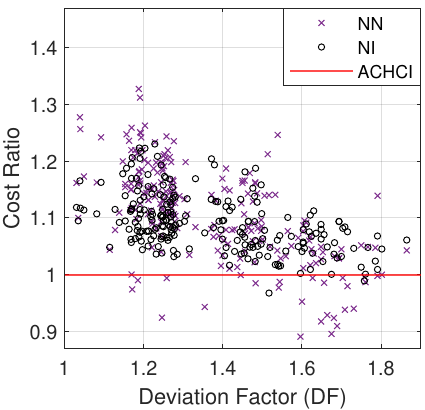}}
        \hspace{0mm}
        % \hspace{0px}
  \subfloat[Cost ratios for various $|V|$\label{H_ratio_points}]{%
        \includegraphics[trim =69mm 103mm 75mm 107mm, clip, width=0.49\linewidth]{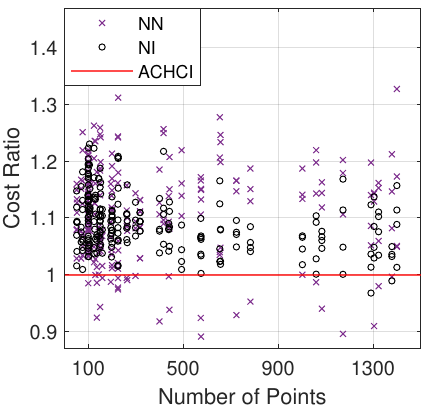}}
  \caption{Ratio of heuristic solution costs to ACHCI cost}
  \vspace{-6mm}
  \label{fig4}
\end{figure}

\begin{figure}[t]
    \centering
    \vspace{-2mm}
        \includegraphics[trim =40mm 103mm 40mm 107mm, clip, width=1\linewidth]{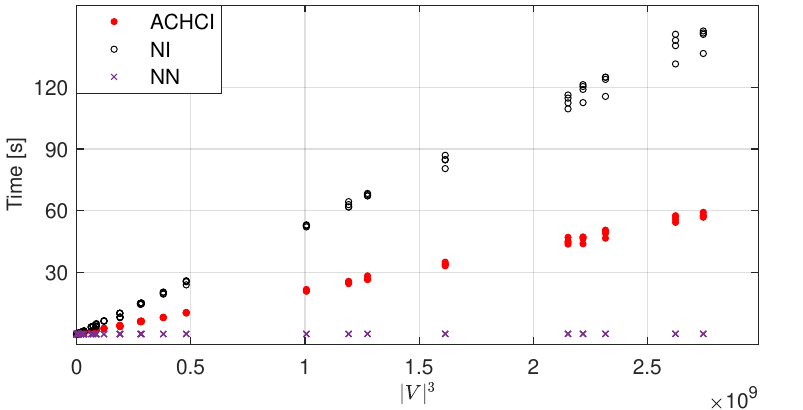}
  \caption{Computation time comparison}
  \label{time} 
\end{figure}

To visualize the effect of DF, the ratio of tour costs obtained using the NN and NI heuristics to that obtained using the ACHCI algorithm is shown in Fig. \ref{H_ratio_DF} for each of the TSPLIB instances studied.
It is clear that the ACHCI algorithm largely outperforms these competing heuristics, indicating that the proposed heuristic is well suited to a variety of non-Euclidean point cloud configurations.
For increasing DF, a diminishing trend is observed in the advantage that the ACHCI algorithm provides, which is attributed to the increasing degree of approximation caused by the MDS when projecting points from an increasingly distorted non-Euclidean space to a Euclidean 2D space.
It is also seen in Fig. \ref{H_ratio_points} that the relative performance of the ACHCI heuristic is largely unaffected by the number of points.
On average, the ACHCI tour cost was 11\% and 9\% lower than the NN and NI heuristics, respectively, outperforming them in 88\% and 99\% of the test cases studied.

\begin{figure}[t]
    \centering
    % \hspace{20px}
  \subfloat[Cost ratio box chart for various DF\label{MH_ratio_DF}]{%
        \includegraphics[trim =69mm 28mm 75mm 28mm, clip, width=0.47\linewidth]{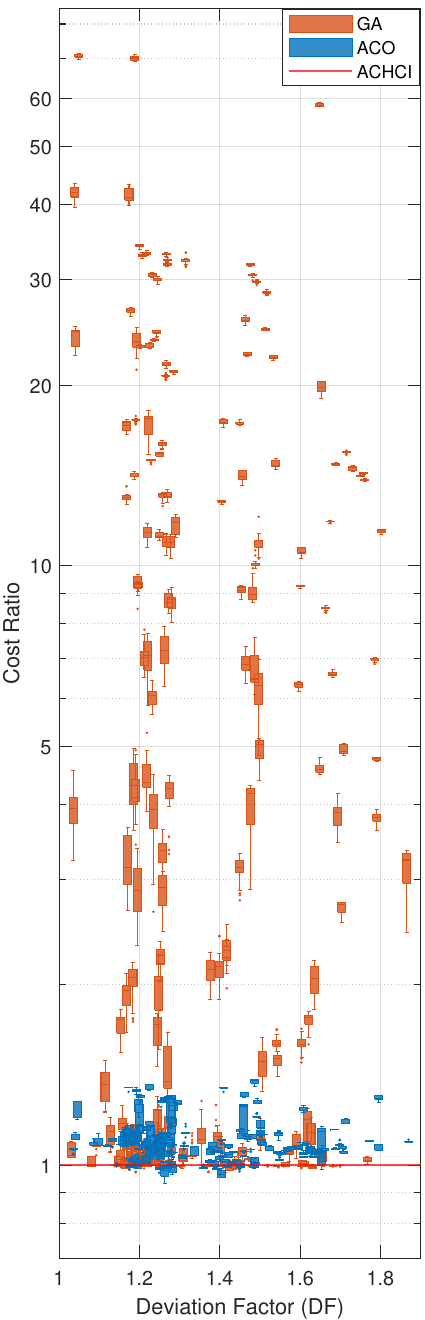}}
        \hspace{0mm}
        % \hspace{0px}
  \subfloat[Cost ratio box chart for various $|V|$\label{MH_ratio_points}]{%
        \includegraphics[trim =69mm 28mm 75mm 28mm, clip, width=0.47\linewidth]{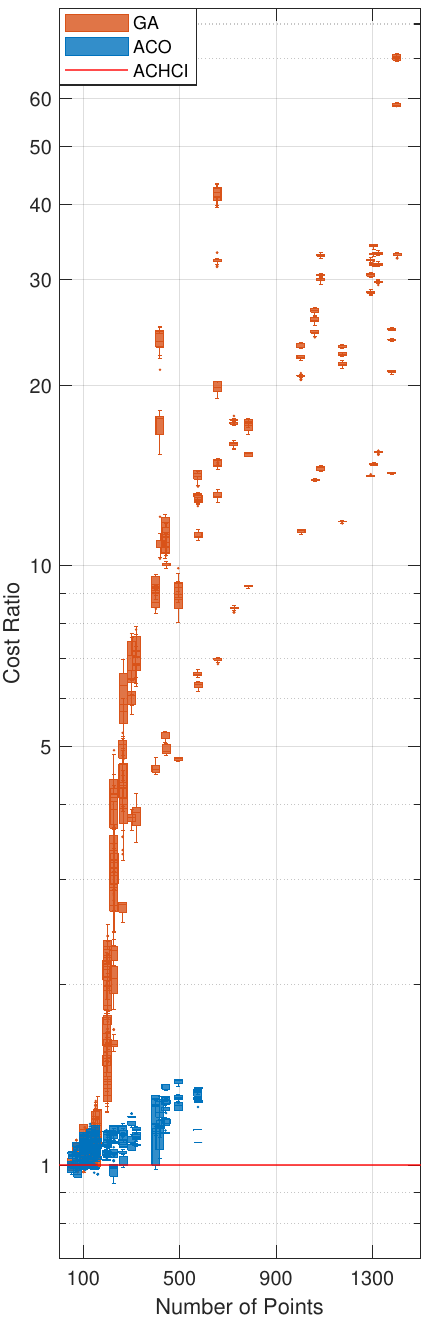}}
  \caption{Ratio of metaheuristic solutions cost to ACHCI cost}
  \label{fig5}
  \vspace{-5mm}
\end{figure}

The ACHCI heuristic exhibits a worst-case time complexity of $O(n^3)$, where $n$ is the number of points in the modified TSPLIB point cloud, as confirmed in Fig.~\ref{time}, where the x-axis shows $n^3$.
This is attributed to the eigenvalue decomposition involved with MDS and the cheapest insertion criteria used when selecting points while building the tour.
The NI heuristic also uses the cheapest insertion criteria but requires more computation time than the ACHCI heuristic because it is not initiated with the convex hull of points, resulting in an increased number of points for insertion.
This results in no solution within the 60-second time limit for larger instances, as seen in Table 2.
Because of the greedy nature of the NN heuristic, tours are produced almost instantaneously regardless of the number of points.

\begin{figure*}[t]
    \centering
    \vspace{2mm}
        \includegraphics[trim =40mm 45mm 35mm 40mm, clip, width=1\linewidth]{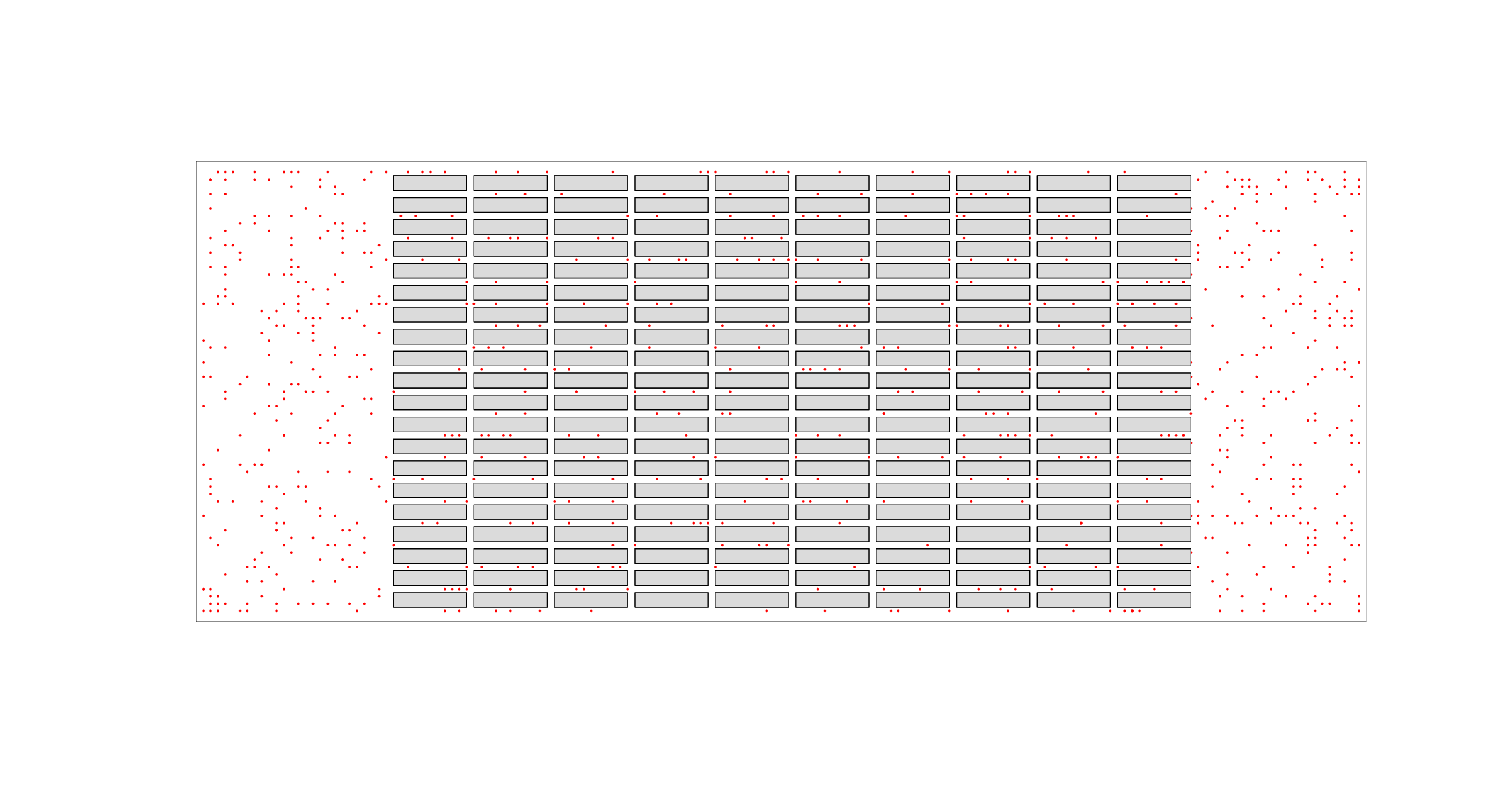}
  \caption{MAPF instance \textit{warehouse-10-20-10-2-1-even-1}}
  \label{MAPF1} 
\end{figure*}

When comparing the performance of metaheuristic approaches, box plots enable the visualization of the stochasticity involved in the final solutions found.
The relative performance of the GA and ACO solutions with respect to the ACHCI solutions decrease with increasing number of points as seen in Fig. \ref{fig5}.
This is because of the exponentially larger search space involved, and while the GA outputs solutions for any number of points because of the random generation of solutions, the ACO does not produce any solutions for larger instances.
This is attributed to the resource intensive operation of constructing each ACO tour by sampling each edge from the initial transition probabilities.
It can be seen however that the performance of GA and ACO for smaller problem sizes is comparable and the relative performance is mostly unaffected by the DF, as seen in Fig. \ref{MH_ratio_DF}.
The ACHCI cost is lower than the mean GA and ACO costs in 87 \% and 95 \% of the studied test cases respectively.
On average, across the tested cases, the \textcolor{black}{ GA tour cost was 8.28 times the corresponding ACHCI cost, demonstrating the poor scalability of GA as the problem size increases.}
Compared to ACO, ACHCI achieved 27\% lower tour costs in instances where ACO found solutions within the computational time budget.

\section{Case Study}\label{caseStudy}

The modified TSPLIB instances studied in Section \ref{computational} made use of radial separators and the $\mathcal{L}_1$ norm to generate a diverse range of problems that were useful for analyzing how ACHCI performance depends on DF and problem size.
However, these synthetic test sets do not directly correlate with the non-Euclidean spaces encountered by autonomous robots deployed in warehouses and flexible manufacturing systems.
To address this limitation, instances from the MAPF benchmark dataset \cite{stern2019mapf} are used as a case study in this section.
The MAPF benchmark provides environment configurations modeled on real-world layouts such as warehouses and city maps.

The first case study is based on the configuration named \textit{warehouse-10-20-10-2-1-even-1} from the MAPF benchmark.
As seen in Fig. \ref{MAPF1}, this layout consists of a $10\times20$ grid of inventory shelves arranged to resemble a large automated warehouse \cite{stern2019mapf}.
A non-Euclidean space is defined by these inventory shelves, preventing straight-line travel and restricting a robot's movement to predefined corridors and intersections.
The robot's objective is to visit all 900 locations of interest, marked by the red dots in Fig. \ref{MAPF1}, while minimizing its NETSP tour cost. 
The second case study uses the \textit{warehouse-10-20-10-2-1-even-21}, which defines 2000 such points to be visited within the same warehouse layout. 

\begin{figure}[t]
    \centering
      \subfloat[ACHCI Tour (Cost: 2828.12)\label{MAPF1E_tours_ACHCI}]{% obtained in 5.62 s
            \includegraphics[trim =40mm 45mm 35mm 40mm, clip, width=1\linewidth]{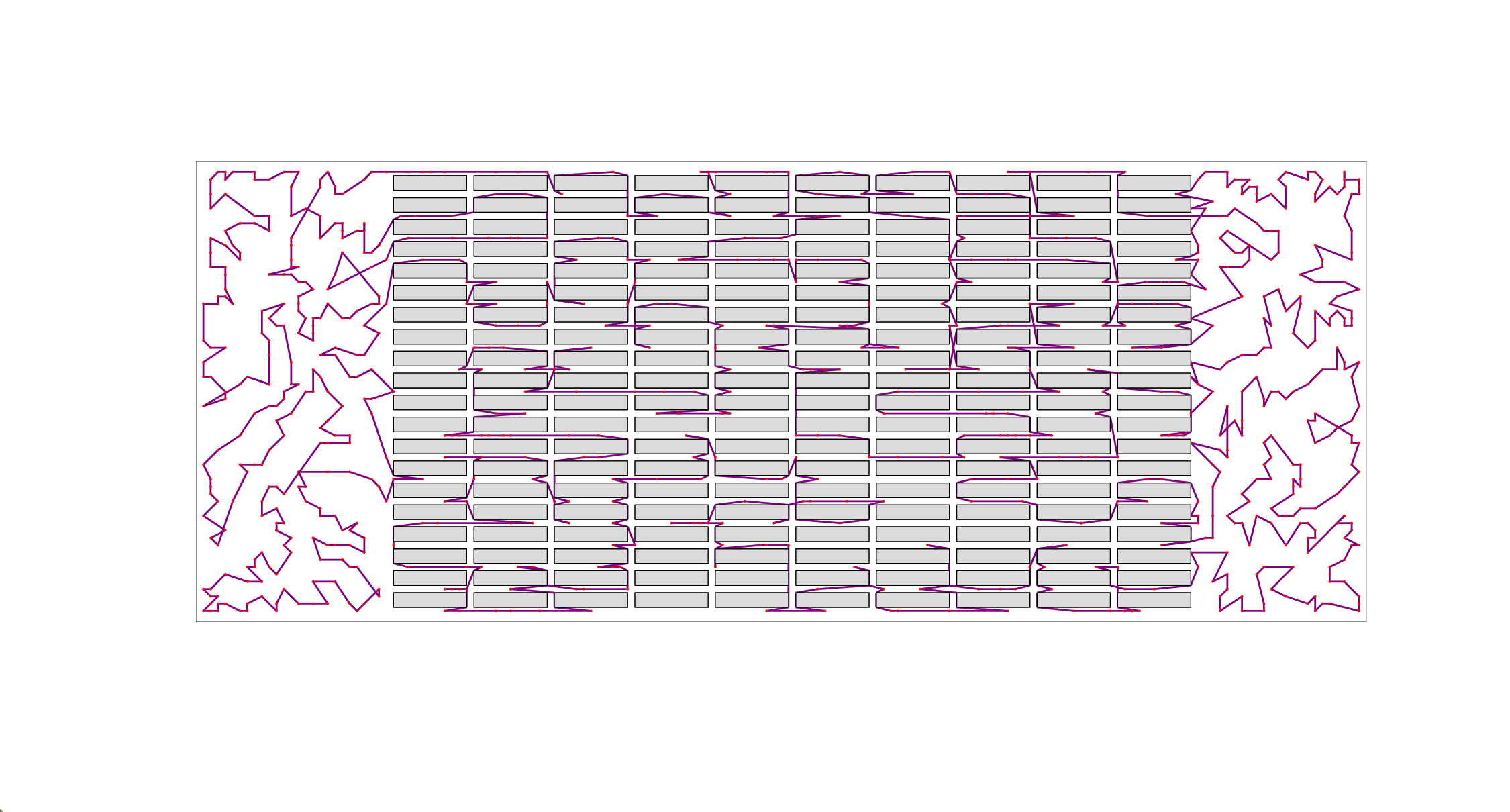}}\\
            \vspace{3mm}
      \subfloat[NN Tour (Cost: 3239.75)\label{MAPF1E_tours_NN}]{% obtained in 0.02 s
            \includegraphics[trim =40mm 45mm 35mm 40mm, clip, width=1\linewidth]{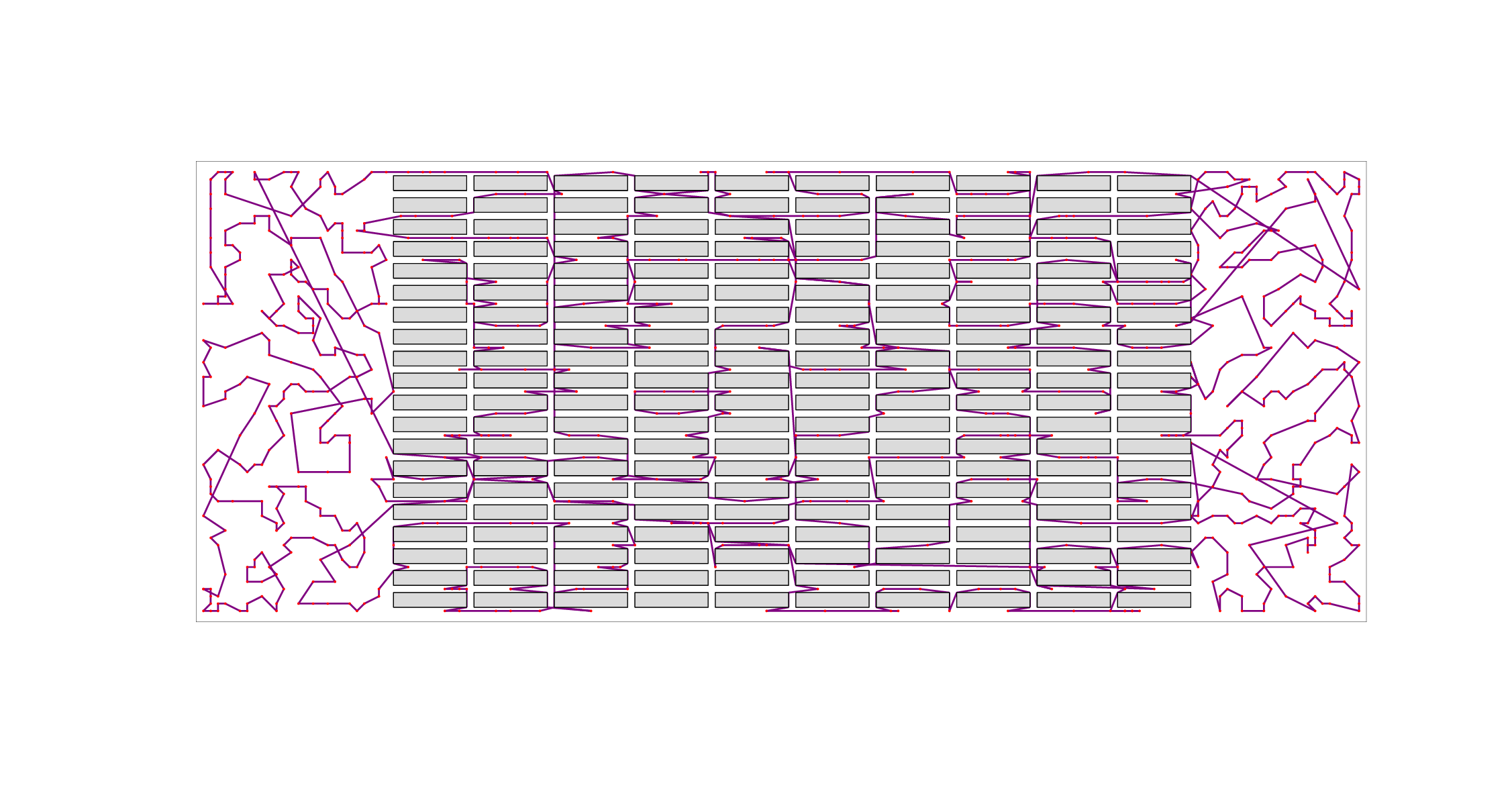}}\\
            \vspace{3mm}
      \subfloat[NI Tour (Cost: 2905.83)\label{MAPF1E_tours_NI}]{%, obtained in 5.21 s
            \includegraphics[trim =40mm 45mm 35mm 40mm, clip, width=1\linewidth]{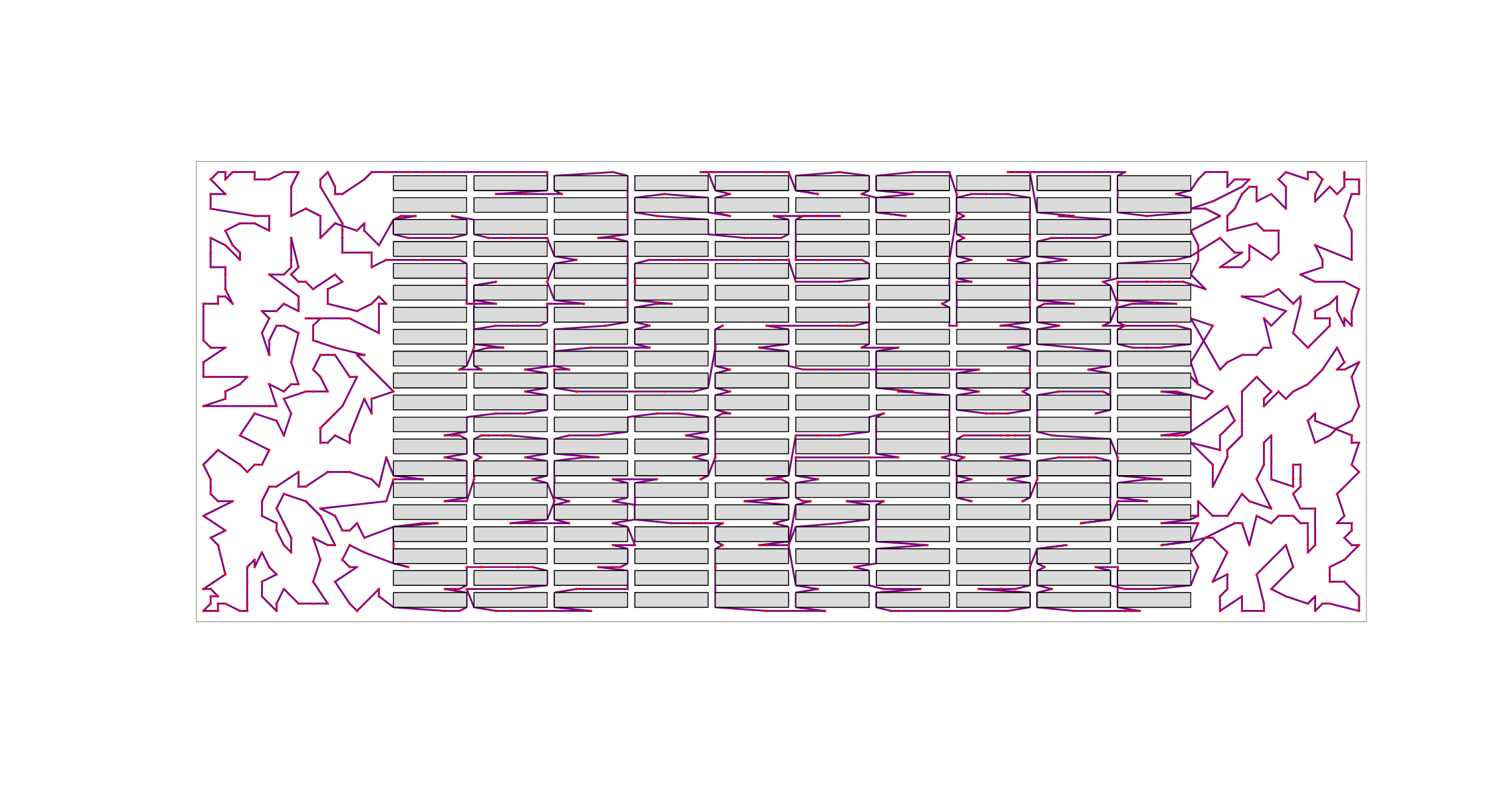}}
\caption{MAPF instance \textit{warehouse-10-20-10-2-1-even-1}}
\label{MAPF1E_tours}
\vspace{-7mm}
\end{figure}

\begin{figure}[t]
    \centering
      \subfloat[ACHCI Tour (Cost: 4033.47)\label{MAPF21E_tours_ACHCI}]{%, obtained in 91.37 s
            \includegraphics[trim =40mm 45mm 35mm 40mm, clip, width=1\linewidth]{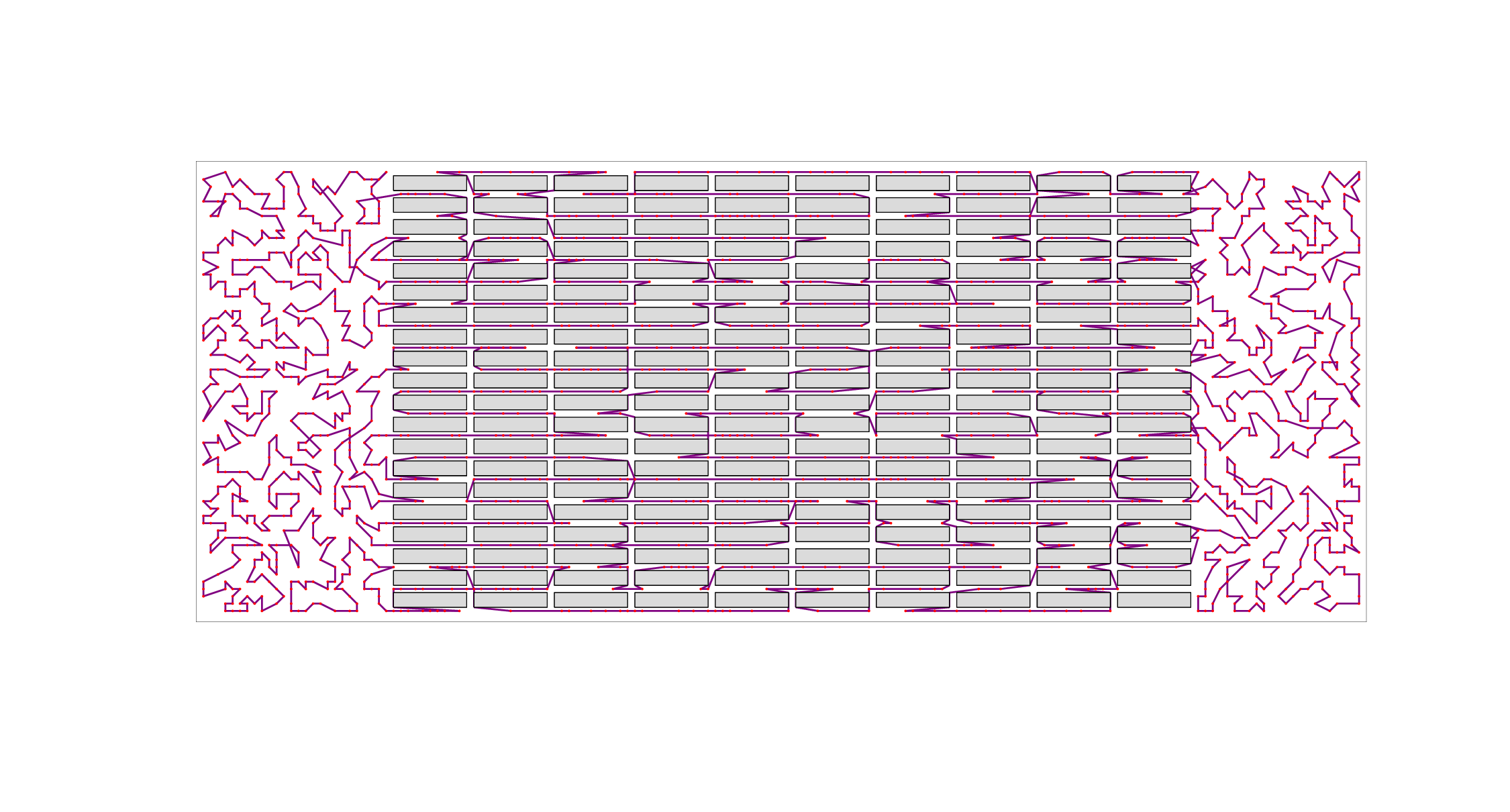}}\\
            \vspace{3mm}
      \subfloat[NN Tour (Cost: 4588.76)\label{MAPF21E_tours_NN}]{%, obtained in 0.09 s
            \includegraphics[trim =40mm 45mm 35mm 40mm, clip, width=1\linewidth]{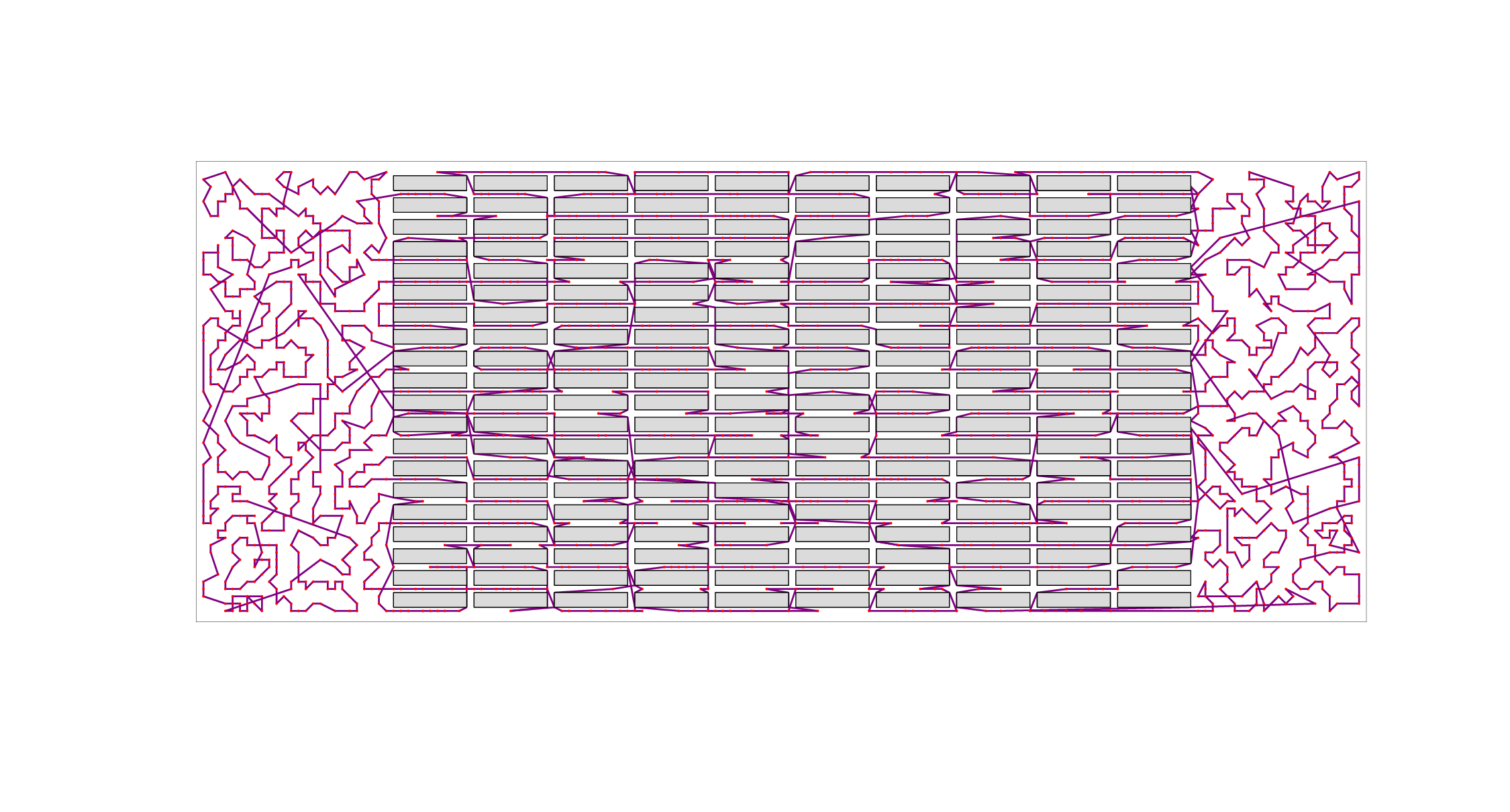}}\\
            \vspace{3mm}
      \subfloat[NI Tour (Cost: 4337.39)\label{MAPF21E_tours_NI}]{%, obtained in 93.17 s
            \includegraphics[trim =40mm 45mm 35mm 40mm, clip, width=1\linewidth]{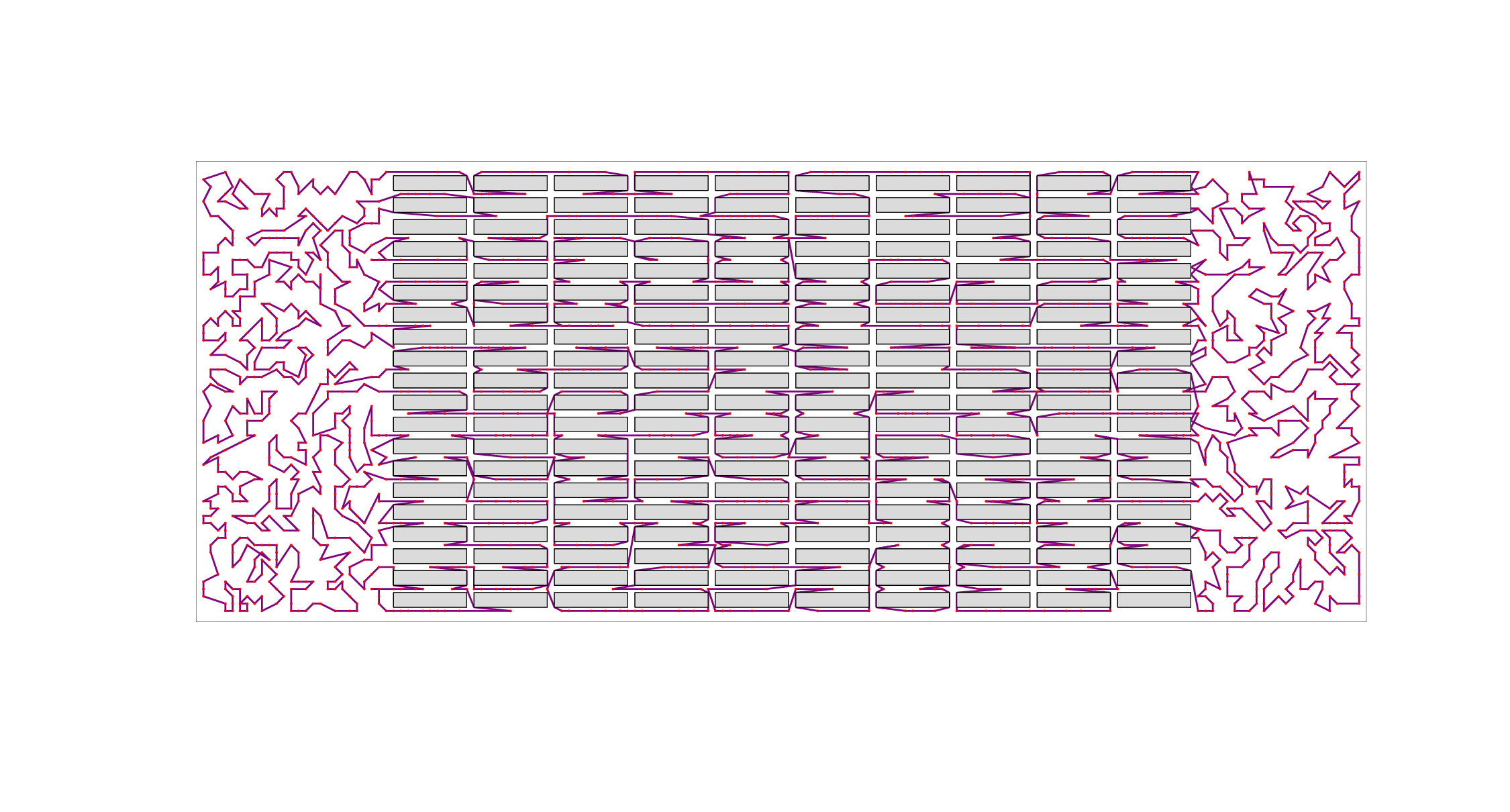}}
\caption{MAPF instance \textit{warehouse-10-20-10-2-1-even-21}}
\label{MAPF21E_tours}
\vspace{-7mm}
\end{figure}

The performance of the ACHCI, NN and NI heuristics is compared in terms of tour costs and computation time, with the  resulting tours and their associated costs provided in Fig. \ref{MAPF1E_tours} and Fig. \ref{MAPF21E_tours}.
In both warehouse scenarios, the ACHCI consistently produces lower cost tours than NN and NI, showing the same trend seen using synthetic test cases of Section \ref{computational}.
Metaheuristic algorithms are excluded from this comparison due to their relatively poor performance under the same computational budget as the heuristics, as well as their high variability in generated tours.

\section{Conclusion}\label{conclusions}

This paper presented the ACHCI algorithm, a lightweight tour construction heuristic for the NETSP,
\textcolor{black}{designed with resource-constrained onboard computation as a target application.}
% tailored to compute NETSP solutions under the constrained computational resources typically available in small form factor robots.
By using MDS to adapt the Euclidean CHCI heuristic for non-Euclidean problems, ACHCI produces lower cost NETSP tours compared to existing heuristic and metaheuristic approaches, while remaining tractable for real-time decentralized execution.
Although the exhaustive computational experiments presented in this paper relied on either using the $\mathcal{L}_1$ norm or adding separators to the TSPLIB benchmark dataset, the same performance was also observed on MAPF warehouse layouts that more closely modeled real-world NETSP problems.

\textcolor{black}{
ACHCI offers a practitioner-friendly alternative to sophisticated heuristics like Christofides and LKH, neither of which has been demonstrated on microcontroller platforms. 
The algorithmic design choices of ACHCI are justified through ablation experiments, whereby MDS is selected for convex hull initialization because it explicitly minimizes global distance distortion, which is critical for identifying  peripheral points. 
Similarly, the cost ratio criterion is empirically shown to yield lower tour costs than the incremental insertion criterion across the majority of tested instances.
}

The \textcolor{black}{reduced tour costs} provide tangible operational advantages by enhancing the capabilities of autonomous robots.
For example, in remote locations, robots deployed with ACHCI can compute routes quickly using onboard processing capabilities without requiring network connectivity for cloud services or high-performance computing, resulting in reliable task completion and reduced idle time.
In drone-based inspections or urban delivery networks, lower tour costs lead to shorter flight times, reduced battery usage, and fewer charging cycles.
\textcolor{black}{
A limitation of this study is that all experiments were conducted on desktop-class systems. 
Future work will evaluate hardware implementations across a range of operational contexts.}

% A limitation of this work is that experiments were completed on desktop-class systems, and therefore, future experimental work will study hardware implementations in various operational contexts.}
% \textcolor{black}{
% Future work will investigate extensions to asymmetric non-metric NETSPs that arise in directed graphs that capture asymmetric turn penalties, one-way road segments or aisle systems.}
% The ACHCI heuristic can be expected to efficiently generate low cost tours for other real-world NETSPs that may arise out of non-Euclidean cost considerations such as fuel consumption, emissions, travel times and even driver comfort. 

\section*{CRediT authorship contribution statement}
\textbf{Mithun Goutham}: Conceptualization, Methodology, Investigation, Software, Visualization, Writing – original draft.
\textcolor{black}{
\textbf{Ethan Rosati}: Investigation, Software.
\textbf{Hollis Schuler}: Investigation, Software.
}
\textbf{Meghna Menon}: Writing – review \& editing, Project administration, Funding acquisition.
\textbf{Sarah Garrow}: Project administration, Funding acquisition.
\textbf{Stephanie Stockar}: Writing – review \& editing, Funding acquisition, Project administration, Supervision.

\section*{Declaration of competing interest} 
The authors declare that they have no known competing financial interests or personal relationships that could have appeared to influence the work reported in this paper.

\section*{Acknowledgements}
This work was supported by Ford Motor Company through the Ford-OSU Alliance Program.

\bibliographystyle{cas-model2-names}
\bibliography{cHull}

\section*{Appendix A. Algorithmic Design Analysis}

\begin{figure*}[!b]
    \centering
    \begin{subfigure}[t]{0.24\textwidth}
        \includegraphics[width=\textwidth, trim=155 242 180 255, clip]{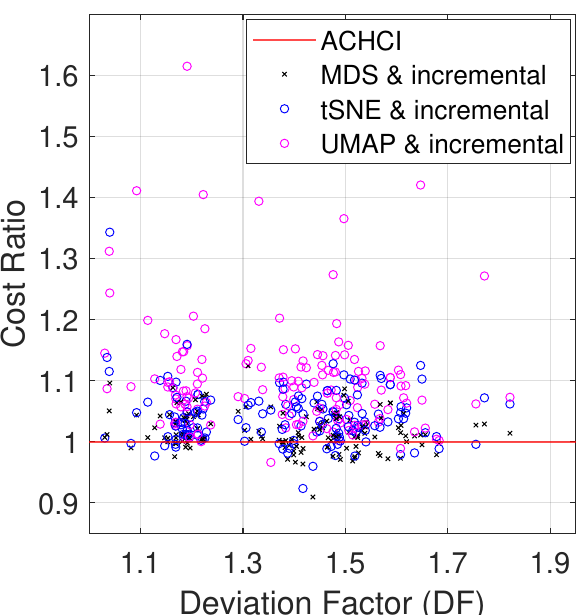}
        \caption{}
        \label{fig:insertion_DF}
    \end{subfigure}
    \hfill
    \begin{subfigure}[t]{0.24\textwidth}
        \includegraphics[width=\textwidth, trim=155 242 180 255, clip]{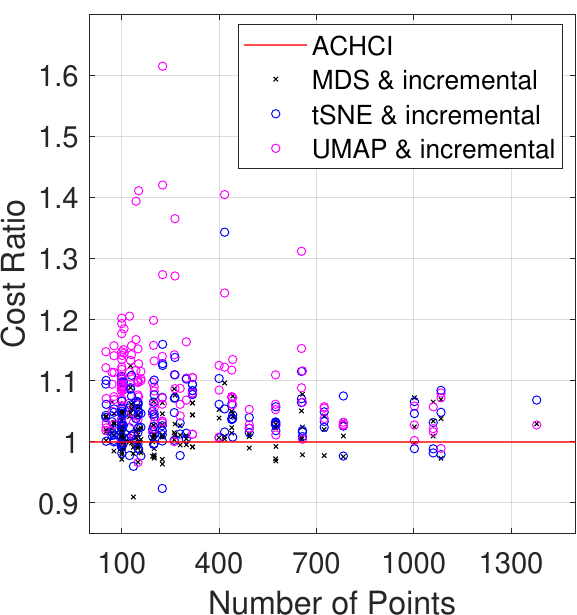}
        \caption{}
        \label{fig:insertion_numPoints}
    \end{subfigure}
    \hfill
    \begin{subfigure}[t]{0.24\textwidth}
        \includegraphics[width=\textwidth, trim=155 242 180 255, clip]{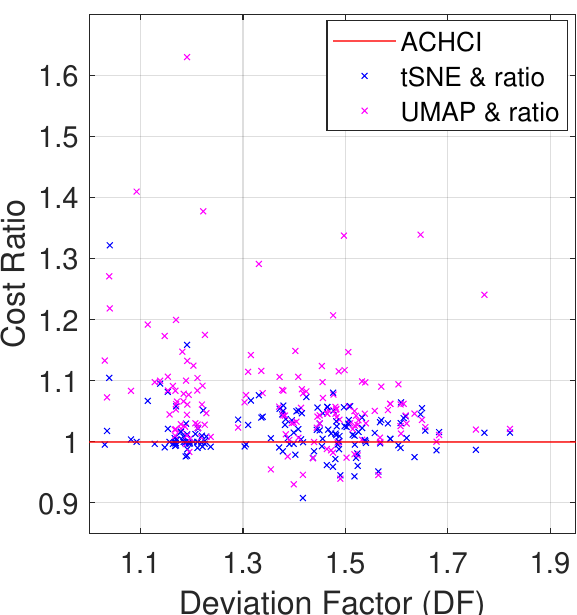}
        \caption{}
        \label{fig:ratio_DF}
    \end{subfigure}
    \hfill
    \begin{subfigure}[t]{0.24\textwidth}
        \includegraphics[width=\textwidth, trim=155 242 180 255, clip]{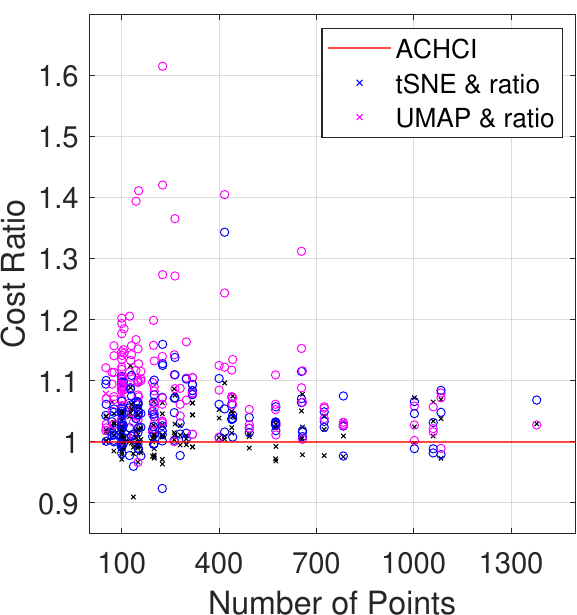}
        \caption{}
        \label{fig:ratio_numPoints}
    \end{subfigure}
    \caption{\textcolor{black}{Cost ratios of initialization method and selection method relative to ACHCI baseline}}
    \label{fig:ablation}
    \vspace{-3mm}
\end{figure*}

\begin{table}[!b]
\centering
\caption{\textcolor{black}{Effect of Insertion Criteria and Embedding on Tour Costs}}
\label{tab:ablation}
\resizebox{\columnwidth}{!}{%
\begin{tabular}{cccc}
\hline
\makecell{Insertion \\ Method} & \makecell{Embedding \\ (convex hull)} & \makecell{Mean Tour Cost \\ Ratio to ACHCI} & \makecell{Variant is better\\ (out of 156)} \\
\hline
\multirow{3}{*}{\makecell{Incremental \\ Insertion}} 
 & MDS & 1.0180 & 40 \\
 & tSNE & 1.0415 & 23 \\
 & UMAP & 1.1000 & 2 \\
\hline
\multirow{3}{*}{\makecell{Insertion \\ Cost Ratio}} 
& MDS$^*$ & 1 & --- \\
& tSNE & 1.0166 & 53 \\
& UMAP & 1.0689 & 18 \\
\hline
\multicolumn{4}{r}{\footnotesize $^*$ Proposed method (ACHCI)} \\
\end{tabular}
}
\vspace{-4mm}
\end{table}

\textcolor{black}{
The effect of the embedding chosen to initialize the convex hull is studied by comparing UMAP and t-SNE as alternatives to the adopted MDS. 
Additionally, the insertion cost ratio criterion $(C_{ik} + C_{kj})/C_{ij}$ used in \textit{Step 3} of the ACHCI algorithm is compared against the incremental insertion cost criterion $C_{ik} + C_{kj}-C_{ij}$.
Performance comparisons similar to Section \ref{computational} are conducted for 39 TSPLIB instances, with non-Euclidean environments created by assigning costs using the $\mathcal{L}_1$ norm or by adding 4, 16 or 32 separators.
}

\textcolor{black}{
UMAP is a nonlinear dimensionality reduction technique using Riemannian geometry and algebraic topology, which models the data as a weighted graph and finds a low-dimensional embedding that preserves it \cite{mcinnes2018umap, healy2024uniform}. 
Unlike MDS, which minimizes a global stress function over all pairwise distances, UMAP balances local neighborhood fidelity with only a degree of global structure preservation by applying attractive forces between neighboring points and repulsive forces between distant ones during optimization.
}

\textcolor{black}{
The UMAP MATLAB implementation (Version 4.6) is used to initiate the convex hull in this study \cite{meehan2021uniform}, with resulting tour costs compared against MDS based tour costs in Fig. \ref{fig:ablation}.
Results using cost ratio based insertions are shown in Fig. \ref{fig:insertion_DF} and \ref{fig:insertion_numPoints}, while those using incremental insertion cost criterion are shown in Fig. \ref{fig:ratio_DF} and \ref{fig:ratio_numPoints}. 
}

\textcolor{black}{
t-SNE is another nonlinear dimensionality reduction technique that converts pairwise similarities between points into probability distributions and minimizes the Kullback-Leibler divergence between the high-dimensional and low-dimensional distributions, prioritizing the preservation of local neighborhood structure over global geometry \cite{van2008visualizing, vandermaaten2014accelerating}. 
Specifically, t-SNE computes a conditional probability $p_{j|i}$ proportional to the Gaussian similarity between points $i$ and $j$ in the high-dimensional space, and a corresponding heavy-tailed Student's t-distribution based probability $q_{ij}$ in the low-dimensional embedding. 
The embedding coordinates are then iteratively optimized using gradient descent to minimize the Kullback-Leibler divergence between the high-dimensional and low-dimensional similarity distributions.
}

\textcolor{black}{
The t-SNE embedding \cite{van2008visualizing,vandermaaten2014accelerating} is computed using the scikit-learn implementation \cite{pedregosa2011scikit} (Version 1.8.0), with \texttt{metric='precomputed'} to directly accept the non-Euclidean cost matrix $C$, setting \texttt{method='exact'} to ensure numerical accuracy. 
Results using cost ratio based insertions are shown in Fig. \ref{fig:insertion_DF} and \ref{fig:insertion_numPoints}, while those using incremental insertion cost criterion are shown in Fig. \ref{fig:ratio_DF} and \ref{fig:ratio_numPoints}. 
}

\textcolor{black}{
As tabulated in Table~\ref{tab:ablation} and shown in Fig.~\ref{fig:ablation}, the proposed ACHCI heuristic, which combines MDS embedding with the insertion cost ratio criterion, produces lower tour costs than the alternative formulations in the majority of test instances. 
This is evidenced by most cost ratio values lying above 1 in Fig.~\ref{fig:ablation}, with the number of instances where the alternative method produces a lower tour cost than ACHCI tabulated in the last column of Table~\ref{tab:ablation}.
The mean tour cost ratios are tabulated in the third column.
}

\textcolor{black}{
These results confirm that MDS, by explicitly minimizing global distance distortion, provides the most reliable convex hull initialization for ACHCI, and that the insertion cost ratio criterion yields lower tour costs relative to the incremental alternative, consistent with the empirical findings of \cite{huang2017investigating} for Euclidean TSPs.
% These findings justify the algorithmic design choices adopted in the proposed ACHCI algorithm for NETSPs.
}

\section*{\textcolor{black}{Appendix B. Time-Limited LKH Comparison}}
\begin{figure*}[t]
    \centering
        \includegraphics[trim =0mm 0mm 0mm 0mm, clip, width=1\linewidth]{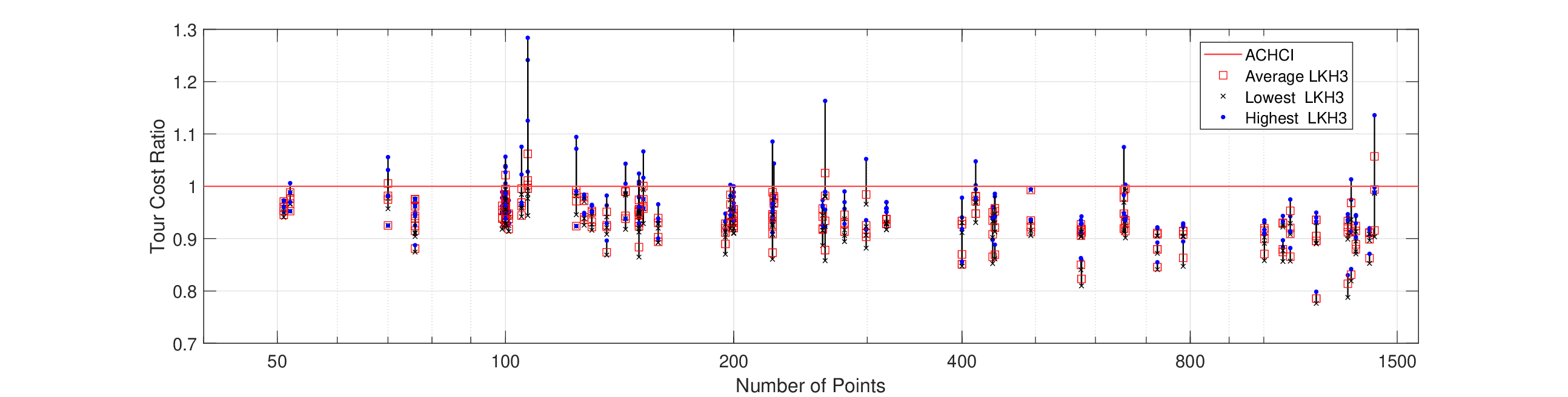}
  \caption{LKH-3 tour cost ratios over 30 repetitions for each NETSP instance}
  \label{LKH-3} 
  \vspace{-2mm}
\end{figure*}

\textcolor{black}{
While Christofides provides a worst-case approximation guarantee for metric TSPs, practical tour quality is generally surpassed by tour-improvement heuristics such as LKH. 
Since neither Christofides nor LKH has been demonstrated on resource-constrained microcontroller platforms, this appendix}
\textcolor{black}{
compares the performance of LKH with ACHCI}
\textcolor{black}{since LKH is a better representative of solutions obtained using centralized high-performance computers.}

\textcolor{black}{
The LKH-3 algorithm \cite{helsgaun2017extension}, an extension of the original Lin-Kernighan-Helsgaun heuristic \cite{helsgaun2000effective}, is widely regarded as one of the most effective TSP heuristics available, consistently producing near-optimal solutions across a range of benchmark instances. 
LKH-3 is typically computed centrally on high-performance hardware, with the resulting tour communicated to the robot for execution, which is incompatible with the decentralized, onboard computation paradigm targeted in this work \cite{lee2022uav, guo2024decentralized, xia2023survey}.
}

\textcolor{black}{
With this context in mind, the} 
\textcolor{black}{performance} 
\textcolor{black}{ of ACHCI is studied in comparison with LKH-3, using its C implementation (Version 3.0.14) maintained at \url{http://webhotel4.ruc.dk/~keld/research/LKH-3/}.}
\textcolor{black}{
Computational experiments are conducted on a desktop equipped with a 2.7 GHz Intel Ultra 9 275HX CPU.}
\textcolor{black}{
The study considers 57 TSPLIB instances, with non-Euclidean environments generated either by assigning edge costs according to the $\mathcal{L}_1$ norm or by introducing 4, 16, or 64 impassable separators.}
\textcolor{black}{
To ensure a fair comparison, ACHCI is also executed in C on the same hardware.
For each instance, the execution time of LKH-3 is limited to the time required by ACHCI to construct its tour, which ranges from 5 ms to 17 s.}

\textcolor{black}{
The ratio of LKH-3 tour costs to ACHCI tour costs is shown in Figure \ref{LKH-3} across all modified TSPLIB instances, plotted against problem size.
Values below the ACHCI baseline indicate instances where LKH-3 produces a lower-cost tour than ACHCI.
Since LKH-3 is inherently stochastic, each NETSP instance is solved 30 times, with the average, lowest and highest LKH-3 costs shown explicitly in Figure \ref{LKH-3}.
The average and minimum LKH-3 costs are generally below the ACHCI baseline, reflecting LKH-3's well-documented solution quality \cite{helsgaun2017extension}. 
}

\textcolor{black}{
LKH-3 exhibits considerable stochastic variability across repetitions, evidenced by the spread between the minimum and maximum costs observed per instance. 
This variability is detrimental in the deployment context targeted by this work, where an autonomous robot must rely on a single-shot computation rather than the mean or the minimum cost obtained after multiple runs\cite{kim2017brain}.
ACHCI being a deterministic construction heuristic produces repeatable solutions without requiring multiple runs, and its cost is comparable with the highly sophisticated LKH-3 algorithm, even outperforming it in a few instances.
}
\textcolor{black}{
A memory comparison is deferred to future work because desktop memory measurements do not reliably translate to microcontroller memory requirements, and no microcontroller implementation of LKH-3 is available.
}

\section*{Appendix C. Supporting Data}
\label{appA: data}

For the sake of brevity, only 14 of the 57 TSPLIB instances are tabulated in Table 2, though Fig. \ref{fig4}, \ref{time} and \ref{fig5} show the results of all 57 instances.

The first column of Table 2 lists the names of the instances, each formatted with an alphabetic prefix followed by a numeric value that indicates the number of points in that instance.
The cause of its non-Euclidean characteristic is listed alongside the instance name, along with the resulting DF.
It can be seen that the DF is positively correlated with the number of separators, though it is also dependent on the spatial distribution of points and their relative position with the added separators.
The remaining columns list the solutions found using the various approaches, with the mean ($\mu$) and standard deviation ($\sigma$) listed for the metaheuristic solutions found at the end of one minute.
Boldfaced entries indicate cases where the competing algorithm produced a lower-cost solution than the ACHCI solution.

\newpage
\noindent  \textbf{Table 2}\\[2pt]
TSP-NE Solution Costs     
\vspace{2mm}

\resizebox{1.9\columnwidth}{!}{
\begin{tabular}{r l|c|c|c|c|c}
\hline
TSPLIB & DF  ~~\textit{(Cause)} & ACHCI & NN & NI & GA $\mu~(\sigma)$ & ACO  $\mu~(\sigma)$\\ \hline
\textit{t70} &1.27  $(\mathcal{L}_1 norm)$  &8.60e+02   & 1.00e+03   & 9.32e+02   & 8.92e+02 (1.77e+01)& 9.26e+02 (3.58e+01)\\ 
                &1.21  $(4~separators)$  &7.69e+02   & 8.86e+02   & 8.88e+02   & 8.17e+02 (6.81e+01)& 7.86e+02 (3.68e+01)\\ 
                &1.44  $(16~separators)$ &1.19e+03   & 1.23e+03   & 1.28e+03   & \textbf{1.19e+03 (1.85e+00)}& \textbf{1.18e+03 (5.96e+00)}\\ 
                &1.60  $(64~separators)$  &2.06e+03   & 2.22e+03   & 2.18e+03   & \textbf{2.06e+03 (4.62e+00)}& 2.16e+03 (3.96e+01)\\ \hline   
\textit{roE100} &1.25  $(\mathcal{L}_1 norm)$  &2.87e+04   & 3.44e+04   & 3.20e+04   & 3.01e+04 (1.92e+03)& 3.09e+04 (3.76e+02)\\ 
                &1.19  $(4~separators)$  &2.36e+04   & 2.79e+04   & 2.88e+04   & 2.59e+04 (4.81e+03)& 2.61e+04 (7.11e+02)\\ 
                &1.39  $(16~separators)$ &3.72e+04   & 3.88e+04   & 4.20e+04   & \textbf{3.71e+04 (8.00e+01)}& \textbf{3.67e+04 (1.49e+02)}\\ 
                &1.61  $(64~separators)$  &6.53e+04   & 7.13e+04   & 7.11e+04   & 6.59e+04 (1.34e+03)& 7.07e+04 (1.28e+03)\\ \hline   
\textit{eil101} &1.28  $(\mathcal{L}_1 norm)$  &8.82e+02   & 9.76e+02   & 9.10e+02   & \textbf{8.82e+02 (4.21e+00)}& 9.50e+02 (1.53e+01)\\ 
                &1.21  $(4~separators)$  &7.22e+02   & 8.65e+02   & 7.59e+02   & 7.34e+02 (3.06e+01)& 7.83e+02 (3.06e+01)\\ 
                &1.48  $(16~separators)$ &1.02e+03   & 1.11e+03   & 1.16e+03   & 1.03e+03 (2.30e+01)& 1.03e+03 (4.35e+01)\\ 
                &1.68  $(64~separators)$  &2.07e+03   & 2.13e+03   & 2.18e+03   & \textbf{2.06e+03 (1.67e+00)}& 2.20e+03 (1.78e+01)\\ \hline   
\textit{bier127} &1.28  $(\mathcal{L}_1 norm)$  &1.60e+05   & 1.80e+05   & 1.81e+05   & 1.68e+05 (7.68e+03)& 1.67e+05 (2.63e+03)\\ 
                &1.16  $(4~separators)$  &1.42e+05   & 1.76e+05   & 1.54e+05   & 1.42e+05 (7.20e+03)& 1.59e+05 (5.69e+03)\\ 
                &1.31  $(16~separators)$ &1.86e+05   & 2.06e+05   & 2.11e+05   & 1.92e+05 (1.68e+04)& 2.11e+05 (5.72e+03)\\ 
                &1.45  $(64~separators)$  &3.80e+05   & 4.27e+05   & 3.93e+05   & 3.85e+05 (6.03e+03)& 4.35e+05 (5.68e+03)\\ \hline   
\textit{roB150} &1.25  $(\mathcal{L}_1 norm)$  &3.53e+04   & 4.19e+04   & 3.83e+04   & 3.97e+04 (6.10e+03)& 3.96e+04 (3.49e+02)\\ 
                &1.17  $(4~separators)$  &3.03e+04   & 3.52e+04   & 3.38e+04   & 3.39e+04 (4.07e+03)& 3.38e+04 (1.01e+03)\\ 
                &1.40  $(16~separators)$ &4.35e+04   & 4.61e+04   & 4.76e+04   & 4.66e+04 (3.44e+03)& 4.49e+04 (8.14e+02)\\ 
                &1.62  $(64~separators)$  &8.31e+04   & 8.81e+04   & 8.94e+04   & 9.73e+04 (1.66e+04)& 8.77e+04 (1.55e+03)\\ \hline   
\textit{u159} &1.24  $(\mathcal{L}_1 norm)$  &5.58e+04   & 5.96e+04   & 6.16e+04   & 6.45e+04 (1.25e+04)& 6.00e+04 (1.81e+03)\\ 
                &1.13  $(4~separators)$  &5.09e+04   & 5.49e+04   & 5.67e+04   & 5.75e+04 (7.10e+03)& 5.60e+04 (1.54e+03)\\ 
                &1.40  $(16~separators)$ &7.22e+04   & 7.33e+04   & 7.79e+04   & 8.02e+04 (8.72e+03)& 7.35e+04 (4.83e+03)\\ 
                &1.63  $(64~separators)$  &1.38e+05   & \textbf{1.37e+05}   & 1.43e+05   & 1.56e+05 (1.95e+04)& 1.42e+05 (7.34e+02)\\ \hline   
\textit{roB200} &1.25  $(\mathcal{L}_1 norm)$  &4.01e+04   & 4.77e+04   & 4.55e+04   & 8.87e+04 (1.35e+04)& 4.52e+04 (3.75e+02)\\ 
                &1.17  $(4~separators)$  &3.41e+04   & 3.80e+04   & 3.80e+04   & 6.56e+04 (1.19e+04)& 3.80e+04 (2.01e+03)\\ 
                &1.40  $(16~separators)$ &4.67e+04   & 5.47e+04   & 5.12e+04   & 9.97e+04 (1.99e+04)& 5.13e+04 (1.31e+03)\\ 
                &1.60  $(64~separators)$  &1.07e+05   & 1.08e+05   & 1.13e+05   & 1.71e+05 (1.55e+04)& 1.11e+05 (1.06e+03)\\ \hline   
\textit{gil262} &1.27  $(\mathcal{L}_1 norm)$  &3.17e+03   & 3.67e+03   & 3.51e+03   & 1.31e+04 (3.14e+03)& 3.65e+03 (1.77e+02)\\ 
                &1.22  $(4~separators)$  &2.87e+03   & 3.20e+03   & 3.05e+03   & 1.27e+04 (2.69e+03)& 3.17e+03 (5.65e+01)\\ 
                &1.50  $(16~separators)$ &3.66e+03   & 3.97e+03   & 4.05e+03   & 1.81e+04 (2.34e+03)& 4.01e+03 (1.34e+02)\\ 
                &1.70  $(64~separators)$  &8.58e+03   & \textbf{8.53e+03}   & 9.28e+03   & 2.31e+04 (1.70e+03)& 9.02e+03 (9.49e+01)\\ \hline   
\textit{in318} &1.26  $(\mathcal{L}_1 norm)$  &5.40e+04   & 6.00e+04   & 5.99e+04   & 3.87e+05 (7.16e+04)& 6.03e+04 (1.13e+03)\\ 
                &1.21  $(4~separators)$  &4.92e+04   & 5.57e+04   & 5.30e+04   & 3.48e+05 (4.80e+04)& 5.66e+04 (7.92e+02)\\ 
                &1.46  $(16~separators)$ &6.42e+04   & 6.96e+04   & 7.02e+04   & 4.44e+05 (3.70e+04)& 7.21e+04 (1.18e+03)\\ 
                &1.69  $(64~separators)$  &1.40e+05   & 1.44e+05   & 1.53e+05   & 5.28e+05 (9.64e+04)& 1.51e+05 (2.32e+03)\\ \hline   
\textit{pcb442} &1.27  $(\mathcal{L}_1 norm)$  &6.28e+04   & 7.58e+04   & 7.08e+04   & 6.87e+05 (4.61e+04)& 8.13e+04 (1.34e+03)\\ 
                &1.22  $(4~separators)$  &5.71e+04   & 6.57e+04   & 6.34e+04   & 6.49e+05 (4.51e+04)& 7.72e+04 (2.30e+03)\\ 
                &1.49  $(16~separators)$ &7.50e+04   & 8.11e+04   & 7.87e+04   & 7.55e+05 (3.18e+04)& 9.56e+04 (2.94e+03)\\ 
                &1.71  $(64~separators)$  &1.77e+05   & \textbf{1.66e+05}   & 1.91e+05   & 8.70e+05 (4.28e+04)& 2.09e+05 (5.41e+03)\\ \hline   
\textit{p654} &1.17  $(\mathcal{L}_1 norm)$  &4.48e+04   & 5.41e+04   & 5.04e+04   & 1.86e+06 (1.50e+05)& -\\ 
                &1.04  $(4~separators)$  &3.98e+04   & 5.09e+04   & 4.64e+04   & 1.66e+06 (1.33e+05)& -\\ 
                &1.32  $(16~separators)$ &5.76e+04   & 5.96e+04   & 6.02e+04   & 1.86e+06 (7.27e+04)& -\\ 
                &1.65  $(64~separators)$  &1.02e+05   & 1.06e+05   & 1.04e+05   & 2.02e+06 (1.25e+05)& -\\ \hline   
\textit{pr1002} &1.27  $(\mathcal{L}_1 norm)$  &3.54e+05   & 4.02e+05   & 3.78e+05   & 7.33e+06 (1.51e+05)& -\\ 
                &1.23  $(4~separators)$  &2.93e+05   & 3.48e+05   & 3.13e+05   & 6.81e+06 (1.80e+05)& -\\ 
                &1.53  $(16~separators)$ &3.53e+05   & 4.03e+05   & 3.59e+05   & 7.85e+06 (1.81e+05)& -\\ 
                &1.80  $(64~separators)$  &7.33e+05   & \textbf{7.33e+05}   & 7.66e+05   & 8.38e+06 (1.52e+05)& -\\ \hline   
\textit{d1291} &1.27  $(\mathcal{L}_1 norm)$  &6.31e+04   & 6.85e+04   & - & 2.04e+06 (3.12e+04)& -\\ 
                &1.23  $(4~separators)$  &6.15e+04   & 7.36e+04   & - & 1.88e+06 (2.83e+04)& -\\ 
                &1.52  $(16~separators)$ &7.45e+04   & 8.50e+04   & - & 2.13e+06 (3.23e+04)& -\\ 
                &1.75  $(64~separators)$  &1.60e+05   & 1.68e+05   & - & 2.25e+06 (1.61e+04)& -\\ \hline   
\textit{fl1400} &1.19  $(\mathcal{L}_1 norm)$  &2.61e+04   & 3.47e+04   & - & 1.83e+06 (3.77e+04)& -\\ 
                &1.05  $(4~separators)$  &2.28e+04   & 2.68e+04   & - & 1.61e+06 (3.30e+04)& -\\ 
                &1.65  $(16~separators)$ &3.41e+04   & 3.58e+04   & - & 2.00e+06 (3.07e+04)& -\\ 
                &2.42  $(64~separators)$  &6.53e+04   & 6.86e+04   & - & 2.15e+06 (3.81e+04)& -\\ \hline    
\end{tabular}}
% \end{table*}

\end{document}